\documentclass[final]{cvpr}

\usepackage{times}
\usepackage{epsfig}
\usepackage{graphicx}
\usepackage{amsmath}
\usepackage{amssymb}
\newtheorem{assumption}{Assumption}
\usepackage{booktabs}
\usepackage{capt-of}
\usepackage{tabularx}
\newcolumntype{C}{>{\centering\arraybackslash}X}
\newcolumntype{L}{>{\raggedright\arraybackslash}X}
\newcolumntype{R}{>{\raggedleft\arraybackslash}X}
\makeatletter
\let\MYcaption\@makecaption
\makeatother
\usepackage[subrefformat=parens]{subcaption}
\makeatletter
\let\@makecaption\MYcaption
\makeatother
\makeatletter
\newcommand\footnoteref[1]{\protected@xdef\@thefnmark{\ref{#1}}\@footnotemark}
\makeatother

\renewcommand{\baselinestretch}{0.94}

\usepackage[pagebackref=true,breaklinks=true,colorlinks,bookmarks=false]{hyperref}

\def\cvprPaperID{10978}
\def\confYear{CVPR 2021}
\begin{document}

\title{\vspace{-0.5em}Unsupervised Learning of Depth and Depth-of-Field Effect from Natural Images\\
  with Aperture Rendering Generative Adversarial Networks\vspace{-1em}}

\author{Takuhiro Kaneko
  \vspace{2mm}\\
  NTT Communication Science Laboratories, NTT Corporation
}

\twocolumn[{
  \renewcommand\twocolumn[1][]{#1}
  \maketitle
  \vspace{-10mm}
  \begin{center}
    \includegraphics[width=\textwidth]{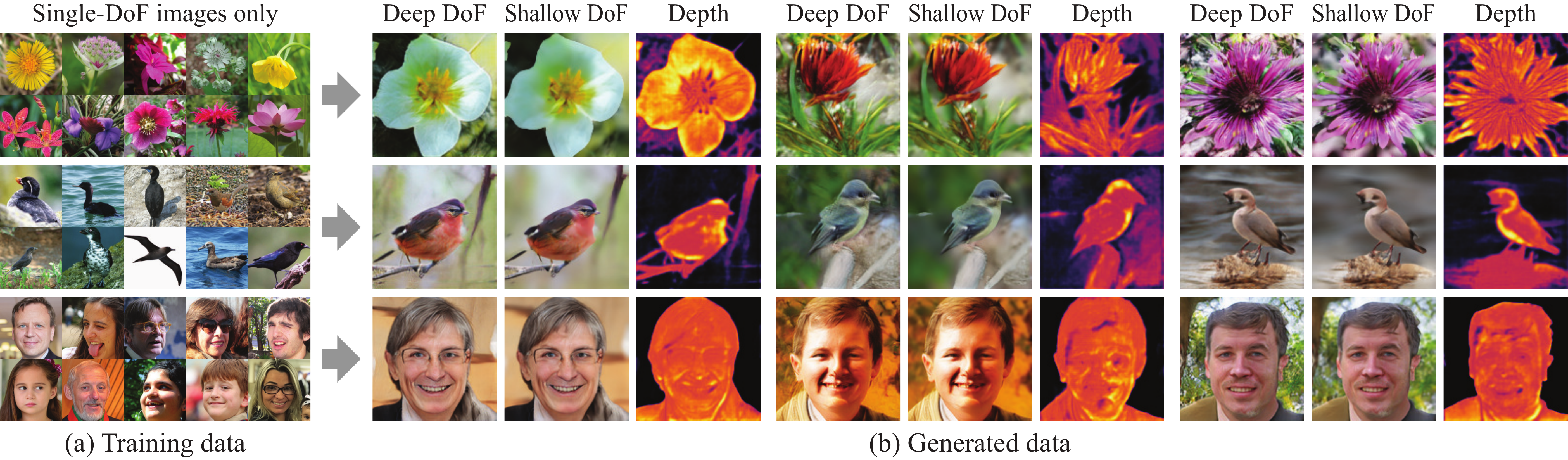}
  \end{center}
  \vspace{-3.5mm}
  \captionof{figure}{\textbf{Unsupervised learning of depth and depth-of-field (DoF) effect from unlabeled natural images.}
    (a) In training, we adopt \textit{only} a collection of single-DoF images \textit{without} any additional supervision (e.g., ground-truth depth, pairs of deep and shallow DoF images, and pretrained model).
    (b) Once trained, our model can synthesize tuples of deep and shallow DoF images and depths from random noise.
    The generated data are beneficial in training a shallow DoF renderer, which also requires \textit{no} external supervision.
    The project page is available at \url{https://www.kecl.ntt.co.jp/people/kaneko.takuhiro/projects/ar-gan/}.}
  \label{fig:concept}
  \vspace{4mm}
}]

\begin{abstract}
  \vspace{-2mm}
  Understanding the 3D world from 2D projected natural images is a fundamental challenge in computer vision and graphics. Recently, an unsupervised learning approach has garnered considerable attention owing to its advantages in data collection. However, to mitigate training limitations, typical methods need to impose assumptions for viewpoint distribution (e.g., a dataset containing various viewpoint images) or object shape (e.g., symmetric objects). These assumptions often restrict applications; for instance, the application to non-rigid objects or images captured from similar viewpoints (e.g., flower or bird images) remains a challenge. To complement these approaches, we propose aperture rendering generative adversarial networks (AR-GANs), which equip aperture rendering on top of GANs, and adopt focus cues to learn the depth and depth-of-field (DoF) effect of unlabeled natural images. To address the ambiguities triggered by unsupervised setting (i.e., ambiguities between smooth texture and out-of-focus blurs, and between foreground and background blurs), we develop DoF mixture learning, which enables the generator to learn real image distribution while generating diverse DoF images. In addition, we devise a center focus prior to guiding the learning direction. In the experiments, we demonstrate the effectiveness of AR-GANs in various datasets, such as flower, bird, and face images, demonstrate their portability by incorporating them into other 3D representation learning GANs, and validate their applicability in shallow DoF rendering.
  \vspace{-1mm}
\end{abstract}
\vspace{-16mm}

\section{Introduction}
\label{sec:introduction}

Natural images are 2D projections of a 3D world.
Addressing the inverse problem, i.e., understanding the 3D world from natural images, is a fundamental challenge in computer vision and graphics.
Owing to its diverse applications in various fields, such as in robotics, content creation, and photo editing, this challenge has been actively studied.

A direct solution to challenge is learning a 3D predictor in a supervised manner using 2D and 3D data pairs or multiview image sets.
However, obtaining such data is often impractical or time-consuming.
To eliminate this process, several studies have attempted to learn 3D representations from single-view images (i.e., with only a single view per training instance).
However, owing to the ill-posed nature, several studies required auxiliary information, such as 2D keypoints~\cite{LTranCVPR2018,AKanazawaECCV2018} or 2D silhouettes~\cite{PHenzlerICCV2019,WChenNeurIPS2019,XLiECCV2020,SGoelECCV2020}, to align object positions or extract a target object from the background.
Other studies required predefined category-specific shape models (e.g., 3DMM~\cite{VBlanzSIGGRAPH1999} and SMPL~\cite{MLoperTOG2015})~\cite{AKanazawaCVPR2018,MWangIJCV2019,BGecerCVPR2019,SSanyalCVPR2019,JShangECCV2020} to obtain clues for reconstruction.
Although they have exhibited promising results, collecting auxiliary information still requires a laborious annotation process, and a shape model requires additional preparation costs and restricts applicable objects.

To eliminate these disadvantages, fully unsupervised learning methods that enable 3D representation learning from single-view images \textit{without} additional supervision and shape models have been devised.
Although this is a severe setting, previous studies have addressed this challenge by imposing assumptions for viewpoint distribution (e.g., a dataset including various viewpoint images)~\cite{TNguyenICCV2019,ASzaboArXiv2019,ANoguchiICLR2020} or object shape (e.g., symmetric objects)~\cite{SWuCVPR2020}.
The first assumption is required to learn 3D representations by sampling diverse viewpoint images.
The second assumption is required to perform stereo reconstruction using a pair of mirrored images.
Although these assumptions are practical for objects of a specific class (e.g., human faces), several objects do not satisfy these assumptions.
For example, these methods are difficult to apply to non-rigid objects or images captured from similar viewpoints (e.g., flower or bird images).

To broaden the application without contradicting previous achievements, in this study, we consider complementary cues inherent in photos that have not been actively used in previous deep generative models (including those above).
In particular, we focus on \textit{focus cues}, in other words, we consider the learning depth\footnote{In this study, we use depth and disparity interchangeably to indicate disparity across a camera aperture.} and the depth-of-field (DoF) effect in the defocus process.
Specifically, instead of imposing an assumption on the \textit{viewpoint distribution}, we do so on the \textit{DoF distribution} (i.e., a dataset including various DoF images), and as shown in Figure~\ref{fig:concept}, we attempt to learn 3D representations (particularly \textit{depth} and \textit{DoF effect}) from a collection of single-DoF images (i.e., images with solely a single DoF setting per training instance).

To achieve this, we propose a novel family of generative adversarial networks (GANs)~\cite{IGoodfellowNIPS2014}, referred to as \textit{aperture rendering GANs (AR-GANs)}, which equip aperture rendering (e.g., light field aperture rendering~\cite{PPSrinivasanCVPR2018}) on top of GANs.
Specifically, AR-GAN initially generates a pair of a deep DoF image and depth from a random noise, and then renders a shallow DoF image from the generated deep DoF image and depth via aperture rendering.
With this mechanism, we can synthesize various DoF images using a virtual camera with an optical constraint on the light field.

When AR-GAN is trained in an unsupervised manner using single-DoF images, two non-trivial challenges are ambiguity between the smooth texture and out-of-focus blurs and ambiguity between the foreground and background blurs, as we cannot obtain explicit supervision of these relationships.
For the first problem, we introduce \textit{DoF mixture learning}, which enables the generator to learn the real image distribution while generating various DoF images.
This learning ensures that the generated images (deep and shallow DoF images) are in real image distribution, and facilitates the learning of the depth, which is a source of connecting deep and shallow DoF images.
For the second problem, based on the observed tendency to focus on the center object when a focused image is considered, we impose a \textit{center focus prior}, which facilitates the focusing of the center while guiding the surroundings to be behind the focal plane.
In practice, we adopt this prior solely at the beginning of training to guide the learning direction.

To evaluate the effectiveness of AR-GAN, we first conducted experiments with comparative and ablation studies on diverse datasets, including flower (Oxford Flowers~\cite{MENilsbackICVGIP2008}), bird (CUB-200-2011~\cite{CWahCUB2002011}), and face (FFHQ~\cite{TKarrasCVPR2019}) datasets.
A significant property of AR-GAN is its portability, which we validated by incorporating AR-GAN into other 3D representation learning GANs (particularly, HoloGAN~\cite{TNguyenICCV2019} and RGBD-GAN~\cite{ANoguchiICLR2020}).
Another significant property of AR-GAN is its ability to synthesize a tuple of deep and shallow DoF images and depth from a random noise, after training.
We utilize this property to train a shallow DoF renderer and empirically demonstrate its utility.

Overall, our contributions are summarized as follows:
\begin{itemize}
  \vspace{-1mm}
  \setlength{\parskip}{1pt}
  \setlength{\itemsep}{1pt}
\item We provide \textit{unsupervised learning of depth and DoF effect from unlabeled natural images}.
  This is noteworthy because it does not impose assumptions on the viewpoint distribution or object shape, which are required in conventional unsupervised 3D representation learning.
\item To achieve this, we propose a novel GAN family (\textit{AR-GANs}), which generate a deep DoF image and depth from a random noise and render a shallow DoF image from them via aperture rendering.
\item To address ambiguities caused by a fully unsupervised setting, we devise \textit{DoF mixture learning} to enable the generator to learn real image distribution using generated diverse DoF images, and develop a \textit{center focus prior} to determine the learning direction.
\item We validate the \textit{effectiveness}, \textit{portability}, and \textit{applicability} of AR-GANs via extensive experiments.
  The project page is available at \url{https://www.kecl.ntt.co.jp/people/kaneko.takuhiro/projects/ar-gan/}.
\end{itemize}

\section{Related work}
\label{sec:related_work}

\noindent\textbf{Generative adversarial networks.}
GANs~\cite{IGoodfellowNIPS2014} have achieved remarkable success in 2D image modeling via a series of advancements (e.g.,~\cite{ABrockICLR2019,TKarrasCVPR2019,TKarrasCVPR2020}).
A substantial property of GANs is their ability to mimic data distribution in a random sampling process without explicitly defining the data distribution.
This allows GANs to learn various distribution types.
For example, recent studies~\cite{XWangECCV2016,JWuNIPS2016,TNguyenICCV2019,PHenzlerICCV2019,ASzaboArXiv2019,ANoguchiICLR2020,YLiaoCVPR2020} have made it possible to learn a 3D-aware image distribution via 3D GAN architectures or 3D representations.
Among them, HoloGAN~\cite{TNguyenICCV2019} and RGBD-GAN~\cite{ANoguchiICLR2020} share a similar motivation with us in terms of learning 3D representations from natural images in a fully unsupervised manner; however, the major difference is that they adopt \textit{viewpoint cues}, whereas we employ \textit{focus cues}.
We empirically demonstrate this difference in Section~\ref{subsec:comparative_study}.
Owing to this difference, the previous and present models are not exchangeable but complementary.
We verify their compatibility in Section~\ref{subsec:portability_analysis} by combining AR-GAN with HoloGAN and RGBD-GAN.

Another related topic is the application of GANs for unsupervised learning of the foreground and background~\cite{CVondrickNIPS2016,JYangICLR2017}.
Although previous and present studies are relevant in terms of learning image compositions, they decompose the image \textit{discretely}, whereas we learn the \textit{continuous depth}.
Furthermore, we can learn the \textit{DoF effect}, which has not been achieved in previous studies.

Other relevant GANs are GANs with measurements~\cite{ABoraICLR2018,APajotICLR2018,SLiICLR2019,TKanekoCVPR2020,TKanekoCVPR2021}, which apply measurements (e.g., mask and noise) before matching a generated image with a real image.
Our aperture rendering functions similarly to those measurements.
However, in the previous work, applicable measurements were limited to those in a \textit{2D image plane}, and effectiveness was solely demonstrated on \textit{synthetically corrupted} images.
By contrast, AR-GAN can learn a DoF effect, which yields a \textit{3D space}.
In the experiments (Section~\ref{sec:experiments}), we verify that this effect can be learned from images taken in \textit{real} scenarios.

\smallskip\noindent\textbf{Unsupervised 3D representation learning.}
As mentioned in Section~\ref{sec:introduction}, the learning of 3D representations from single-view images has garnered attention owing to its data collection advantage.
To address this challenge, several studies have employed auxiliary information as clues for reconstruction, such as 2D keypoints~\cite{LTranCVPR2018,AKanazawaECCV2018}, 2D silhouettes~\cite{PHenzlerICCV2019,WChenNeurIPS2019,XLiECCV2020,SGoelECCV2020}, or shape models~\cite{AKanazawaCVPR2018,MWangIJCV2019,BGecerCVPR2019,SSanyalCVPR2019,JShangECCV2020}.
By contrast, we attempt to address this challenge with \textit{no} additional supervision and \textit{no} predefined model to reduce costs from laborious annotation and model preparation.

Recently, some studies~~\cite{TNguyenICCV2019,ASzaboArXiv2019,ANoguchiICLR2020,SWuCVPR2020} have addressed this; however, their assumptions and objectives differ from ours.
They impose assumptions for the \textit{viewpoint distribution} or \textit{object shape}, whereas we impose an assumption for the \textit{DoF distribution}.
Owing to this assumption difference, they can learn \textit{3D meshes}~\cite{ASzaboArXiv2019}, \textit{depth}~\cite{ANoguchiICLR2020,SWuCVPR2020}, \textit{albedo}~\cite{SWuCVPR2020}, \textit{texture}~\cite{ASzaboArXiv2019}, \textit{light}~\cite{SWuCVPR2020}, and \textit{viewpoints}~\cite{TNguyenICCV2019,ASzaboArXiv2019,ANoguchiICLR2020,SWuCVPR2020}, whereas AR-GAN can learn the \textit{depth} and \textit{DoF effect}.
Therefore, AR-GAN can be considered a model that can complement (not replace) previous models.
We validate this statement in Section~\ref{subsec:portability_analysis} by incorporating AR-GAN into HoloGAN~\cite{TNguyenICCV2019} and RGBD-GAN~\cite{ANoguchiICLR2020}.

\smallskip\noindent\textbf{Monocular depth estimation.}
Monocular depth estimation involves predicting the depth when a single image is given.
A successful approach involves training a depth predictor using paired or consecutive data, such as image and depth pairs~\cite{DEigenNIPS2014,FLiuTPAMI2015,ILaina3DV2016,YKuznietsovCVPR2017,DXuCVPR2017,HFuCVPR2018}, stereo pairs~\cite{RGargECCV2016,CGodardCVPR2017,KXianCVPR2020}, and videos~\cite{TZhouCVPR2017,ZYinCVPR2018,CWangCVPR2018}.
Although this approach is a promising solution, collecting such data is often impractical or time-consuming.

In another direction, some studies~\cite{PPSrinivasanCVPR2018,SGurCVPR2019} have proposed the adoption of focused and all-in-focus image pairs, including learning the depth in the process of reconstructing the focused image from an all-in-focus image.
Although this study is inspired by their success, the main difference is that they require paired supervision between focused and all-in-focus images, whereas ours does not need it.
However, owing to this difference, our task is very challenging; therefore, in this study, we did not attempt to achieve high-quality depth estimation comparable to supervised methods.
Instead, in the experiments, we compared AR-GAN with a previous fully unsupervised depth estimation model (i.e., RGBD-GAN~\cite{ANoguchiICLR2020}) and demonstrated the utility of AR-GAN in this challenging setting (Section~\ref{subsec:comparative_study}).

\smallskip\noindent\textbf{DoF rendering.}
The DoF or Bokeh effect is a popular photography technique, and its synthesis has garnered considerable interest in computer vision and graphics.
To achieve this without prior knowledge of geometry and lightning, previous studies adopted stereo pairs~\cite{JTBarronCVPR2015}, a stack of images taken in different camera settings~\cite{DEJacobsSCTLTR2012,SWHasinoffICCV2007}, and a segmentation mask~\cite{XShenECCV2016,NWadhwaTOG2018} to determine the degree of blur.
Although they have exhibited remarkable results, they are limited owing to their general dependence on a manually defined DoF renderer.
To address this limitation, end-to-end supervised learning methods~\cite{PPSrinivasanCVPR2018,LWangTOG2018,AIgnatovCVPRW2020,MQianArXiv2020}, which train a DoF renderer using pairs of shallow and deep DoF images, were devised.
Recently, an unpaired learning method~\cite{JYZhuICCV2017} was also proposed.
This method eliminates the requirement for paired supervision; however, set-level supervision (i.e., supervision of whether each image is a deep or shallow DoF image) remains necessary.
By contrast, we focus on training a DoF renderer in a \textit{fully} unsupervised manner.
We demonstrate the effectiveness of our approach in Section~\ref{subsec:application}.

\section{Preliminaries}
\label{sec:preliminaries}

\subsection{GANs}
\label{subsec:gan}

We briefly introduce two previous works on which our model is based.
The first is GAN~\cite{IGoodfellowNIPS2014}, which learns data distribution using the following objective:
\begin{flalign}
  \label{eqn:gan}
  {\cal L}_{\text{GAN}}
  = & \: \mathbb{E}_{I^r \sim p^r(I)} [ \log C(I^r) ]
  \nonumber \\
  + & \: \mathbb{E}_{z \sim p(z)} [ \log (1 - C(G(z))) ],
\end{flalign}
where, given a random noise $z$, a generator $G$ generates an image $I^g = G(z)$ that can deceive a discriminator $C$ by minimizing this objective, whereas $C$ distinguishes $I^g$ from a real image $I^r$ by maximizing this objective.
Here, superscripts $r$ and $g$ denote the real and generated data, respectively.
Using this min-max game, the generative distribution $p^g(I)$ approaches the real distribution $p^r(I)$.

\subsection{Light field aperture rendering}
\label{subsec:lf_aperture_rendering}

Light field aperture rendering~\cite{PPSrinivasanCVPR2018} is a type of differentiable aperture rendering.\footnote{Another representative aperture rendering is compositional aperture rendering~\cite{PPSrinivasanCVPR2018}, which discretely models disparities using a stack of blur kernels.
  In the initial experiments, we determined that light field aperture rendering, which models the light field within a camera, is more compatible with our unsupervised learning.
  This is possibly because the learning clues are few in our unsupervised learning; therefore, an explicit camera constraint via light field aperture rendering works sufficiently.}
Its objective is to learn an aperture renderer $R$ that synthesizes a shallow DoF image $I_s(\mathbf{x}) = R(I_d(\mathbf{x}), D(\mathbf{x}))$, given a deep DoF image $I_d (\mathbf{x})$ and depth $D(\mathbf{x})$.\footnote{In the original study~\cite{PPSrinivasanCVPR2018}, $D(\mathbf{x})$ is estimated from $I(\mathbf{x})$.
  However, this estimation is not adopted in AR-GAN; hence, we omitted it here.}
Here, $\mathbf{x}$ represents the spatial coordinates of the light field on the image plane.
When $I_d(\mathbf{x})$ is directly warped into the viewpoint in the light field based on $D(\mathbf{x})$, holes can appear in the resulting light field.
Instead, a trainable neural network $T$ is adopted to expand $D(\mathbf{x})$ into a depth map $M(\mathbf{x}, \mathbf{u})$ for each view in the light field:
\begin{flalign}
  \label{eqn:depth_map}
  M(\mathbf{x}, \mathbf{u}) = T(D(\mathbf{x})),
\end{flalign}
where $\mathbf{u}$ denotes the angular coordinates of the light field on the aperture plane.
Subsequently, $I_d(\mathbf{x})$ is warped into each view in the light field using the depth map $M(\mathbf{x}, \mathbf{u})$:\footnote{The depth of the focal plane can be learned explicitly by adding the parameterized offset $\hat{m}$ to $M$ in Equation~\ref{eqn:light_field}.
  However, we do not do so under the assumption that it is determined per image $I_d$ and internally represented and optimized in $D$, which is used in Equation~\ref{eqn:depth_map}.
  In this case, the focal plane exists at $D = 0$, while out-of-focus occurs in $|D| > 0$.}
\begin{flalign}
  \label{eqn:light_field}
  L(\mathbf{x}, \mathbf{u}) = I_d(\mathbf{x} + \mathbf{u} M(\mathbf{x}, \mathbf{u})),
\end{flalign}
where $L(\mathbf{x}, \mathbf{u})$ is the simulated camera light field.
Finally, it is integrated to render a shallow DoF image $I_s(\mathbf{x})$:
\begin{flalign}
  \label{eqn:shallow_dof}
  I_s(\mathbf{x}) = \sum_{\mathbf{u}} A(\mathbf{u}) L(\mathbf{x}, \mathbf{u}),
\end{flalign}
where $A(\mathbf{u})$ is an indicator that represents the disk-shaped camera aperture.
Hereafter, for simplicity, we omit $\mathbf{x}$ and $\mathbf{u}$ when they are not required.

\section{Aperture rendering GANs: AR-GANs}
\label{sec:ar-gan}

\subsection{Problem statement}
\label{subsec:problem_statement}

We begin by defining the problem statement.
We consider a fully unsupervised setting in which we cannot obtain any supervision or pretrained model except for an image collection.
As discussed in Section~\ref{sec:related_work}, typical end-to-end focus-based monocular depth estimation methods (e.g., \cite{PPSrinivasanCVPR2018,SGurCVPR2019}), and DoF rendering methods (e.g., \cite{PPSrinivasanCVPR2018,AIgnatovCVPRW2020,MQianArXiv2020,JYZhuICCV2017}) achieve their objectives using a \textit{conditional} model (i.e., a deep DoF image is used as the input, and a depth or shallow DoF image is estimated based on it).
However, in our fully unsupervised setting, we cannot employ this formulation as we cannot obtain either ground-truth depth or supervision of whether each image is a deep or shallow DoF image.

Alternatively, we aim to learn an \textit{unconditional} generator $G(z)$ that can generate a tuple of a deep DoF image, depth, and shallow DoF image, i.e., $(I_d^g, D^g, I_s^g)$, from a random noise $z$.
When the training images are extremely biased in terms of the DoF (e.g., all images are all-in-focus), it is difficult to obtain focus cues from the images; hence, we impose the following assumption on an image distribution:
\begin{assumption}
  \label{assumption:ar-gan}
  The DoF setting is different for each image, and the dataset includes various DoF images.
\end{assumption}
Note that we do not have to collect a \textit{set/pair} of various DoF images for each training instance.
We observed that this assumption is satisfied by typical natural image datasets (e.g., flower~\cite{MENilsbackICVGIP2008}, bird~\cite{CWahCUB2002011}, and face~\cite{TKarrasCVPR2019} datasets shown in Figure~\ref{fig:concept}).
Under this assumption, we aim to learn the above-mentioned generator in a \textit{wisdom of crowds} approach.

\begin{figure}[t]
  \centering
  \includegraphics[width=\columnwidth]{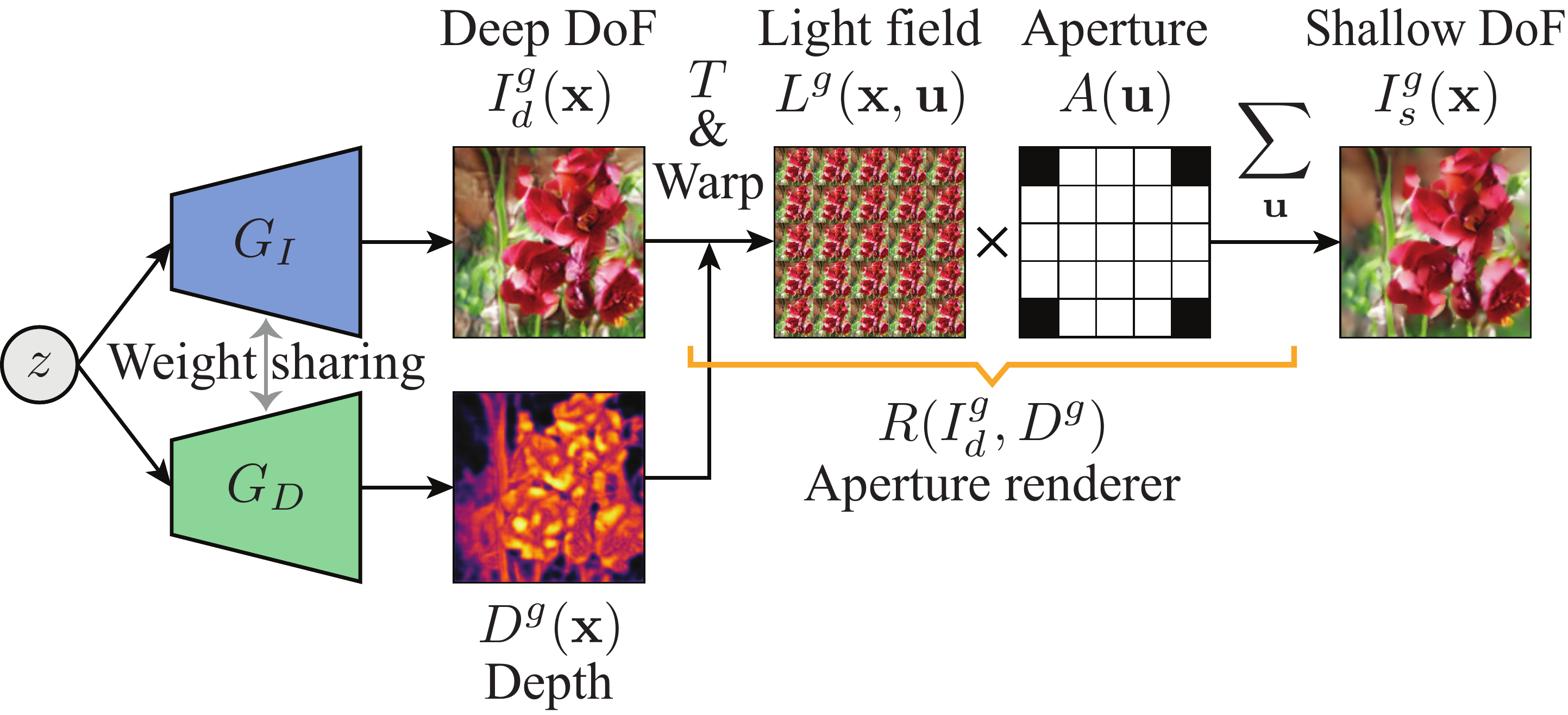}
  \caption{\textbf{Overall pipeline of AR-GAN generator.}
    The AR-GAN generator first generates a deep DoF image $I_d^g$ and depth $D^g$ from a random noise $z$, and then renders a shallow DoF image $I_s^g$ from $I_d^g$ and $D^g$ using the aperture renderer $R$.}
  \label{fig:argan_generator}
  \vspace{-2mm}
\end{figure}

\subsection{Overall pipeline}
\label{subsec:overall_pipeline}

The overall pipeline of the AR-GAN generator is illustrated in Figure~\ref{fig:argan_generator}.
When given a random noise $z$, we first generate a deep DoF image $I_d^g$ and depth $D^g$ as follows:
\begin{flalign}
  I_d^g = G_I(z), D^g = G_D(z).
\end{flalign}
In practice, we share the weights between $G_I$ and $G_D$ except for the last layer because the image and depth exhibit high correlation.
A previous study~\cite{MYLiuNIPS2016} demonstrated that this kind of weight sharing is beneficial in learning a joint distribution between relevant domains.
Subsequently, we render a shallow DoF image $I_s^g$ from the generated $I_d^g$ and $D^g$ using the aperture renderer $R$ described in Section~\ref{subsec:lf_aperture_rendering}.

Typical GANs apply a discriminator $C$ to the final output of the generator (i.e., $I_s^g$ in our case).
However, in AR-GAN, both generators (i.e., $G_I$ and $G_D$) and $R$ are trainable.
Hence, without constraints, they could compete for roles.
For example, they can drift into an extreme solution (e.g., $R$ learns strong out-of-focus, and $G_I$ learns an over-deblurred image).
To address this, we develop \textit{DoF mixture learning}, which is detailed in the next section.

\subsection{DoF mixture learning}
\label{subsec:dof_mixture_learning}

A possible solution to this problem is regularizing $G_I$ using an explicit distance metric (e.g., L1, L2, or perceptual loss~\cite{JJohnsonECCV2016,ADosovitskiyNIPS2016}) such that $I_d^g$ is approximate to $I_s^g$.
However, this solution disrupts the depth learning (Section~\ref{subsubsec:ablation_study_dof_mixture_learning}).

\begin{figure}[t]
  \centering
  \includegraphics[width=\columnwidth]{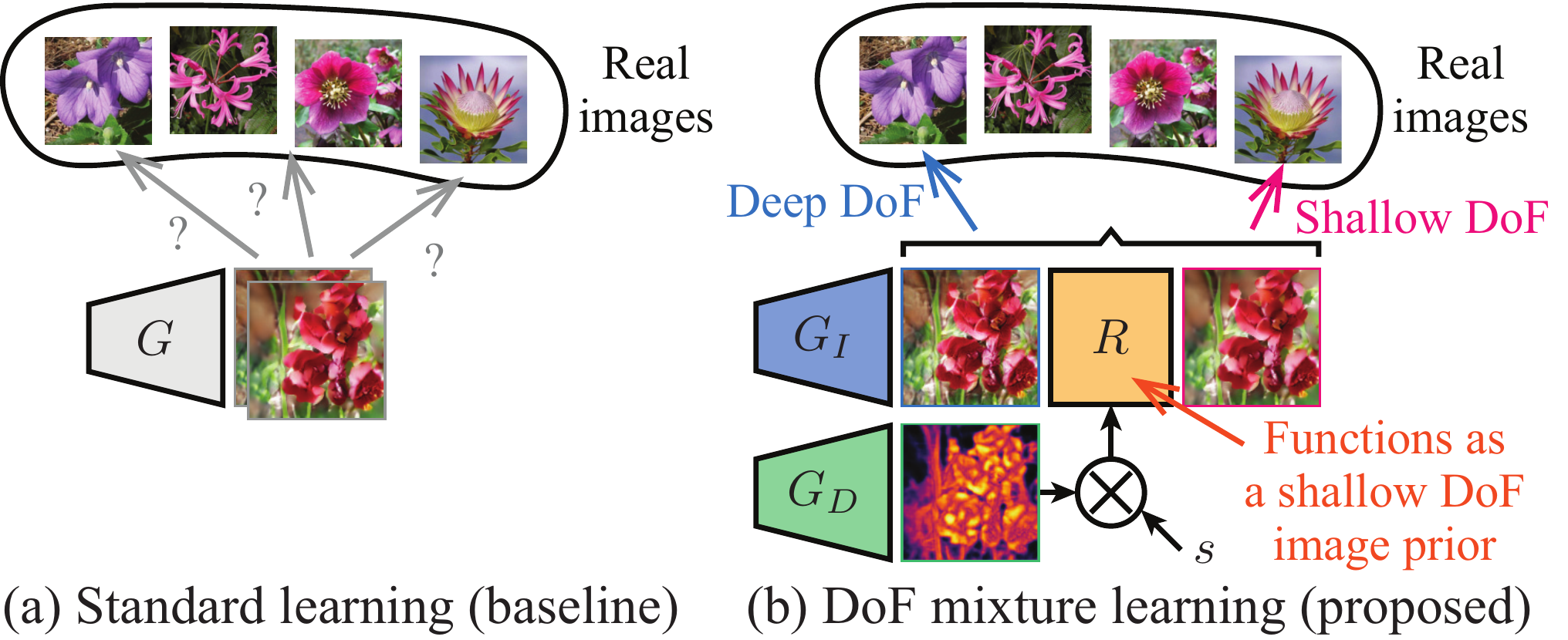}
  \caption{\textbf{Comparison of standard and DoF mixture learning.}}
  \label{fig:dof_mixture_learning}
  \vspace{-2mm}
\end{figure}

Alternatively, we introduce \textit{DoF mixture learning}.
Figure~\ref{fig:dof_mixture_learning} illustrates the comparison between standard and DoF mixture learning.
In standard GAN training, the generator attempts to cover the real image distribution using images without constraints.
By contrast, in the DoF mixture learning, the generator attempts to represent the real image distribution using diverse DoF images whose extent is adjusted by a scale factor $s$.
More precisely, in our AR-GAN, the GAN objective (Equation~\ref{eqn:gan}) is rewritten as follows:
\begin{flalign}
  \label{eqn:ar-gan}
  & {\cal L}_{\text{AR-GAN}}
  = \mathbb{E}_{I^r \sim p^r(I)} [ \log C(I^r) ]
  \nonumber \\
  & \:\: + \mathbb{E}_{z \sim p(z), s \sim p(s)} [ \log (1 - C(R(G_I(z), s G_D(z)))) ],
\end{flalign}
where $s \in [0, 1]$; when $s = 0$, a deep DoF image (almost equal to $I_d^g$) is rendered, whereas when $s = 1$, a shallow DoF image ($I_s^g$) is rendered.
Intuitively, the aperture renderer $R$, which has an optical constraint on the light field, functions as a shallow DoF image prior.
Under Assumption~\ref{assumption:ar-gan} (a real image distribution $p^r(I)$ includes both deep $I_d^r$ and shallow $I_s^r$ DoF images), this prior encourages the generated deep $I_d^g$ and shallow $I_s^g$ DoF images to approximate $I_d^r$ and $I_s^r$, respectively.
This also facilitates the learning of $D^g$, which is a source of the $I_d^g$ and $I_s^g$ connection.

\begin{figure}[b]
  \vspace{-2mm}
  \centering
  \includegraphics[width=0.9\columnwidth]{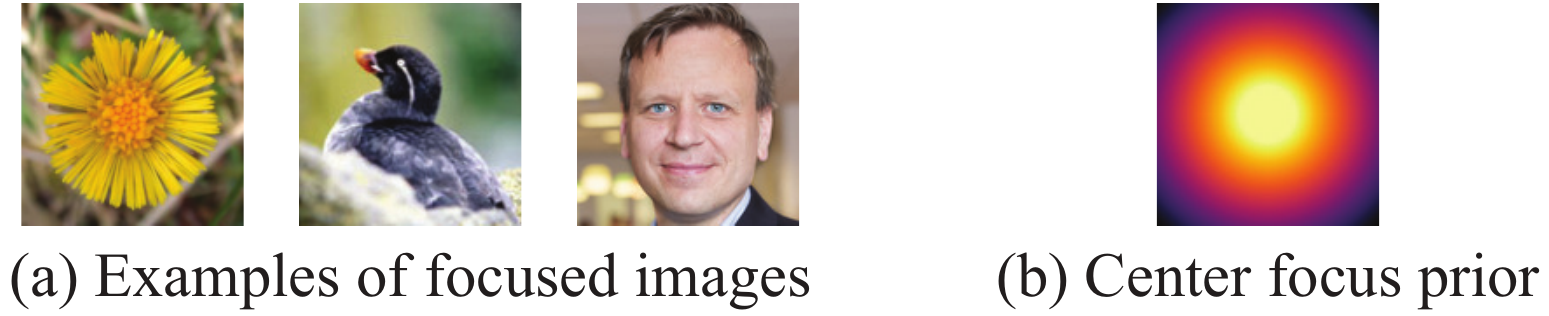}
  \caption{\textbf{Examples of focused images and center focus prior.}
    In (b), light color indicates the foreground.}
  \label{fig:center_focus_prior}
\end{figure}

In practice, we determined that sampling $s$ from a binomial distribution, i.e., $p(s) = B(1, p_s)$, works optimally, where $p_s$ indicates a probability of $s = 1$.
In Section~\ref{subsec:ablation_study}, we examine the effect of the $p_s$ value.
It was manually determined for simplicity; however, optimizing it in a data-driven approach is a potential direction for future work.

\subsection{Center focus prior}
\label{subsec:center_focus_prior}

Another challenge unique to unsupervised depth and DoF effect learning is to the difficulty in distinguishing foreground and background blurs without any constraint or prior knowledge.
Although not all images satisfy this, focused images tend to be captured when the main targets are positioned at the center, as shown in Figure~\ref{fig:center_focus_prior}(a).
Based on this observation, we impose a center focus prior defined by
\begin{flalign}
  \label{eqn:center_focus_prior}
  D_p = \begin{cases}
    0 & (r <= r_{\text{th}}) \\
    -g \cdot (r - r_{\text{th}}) & (r > r_{\text{th}}),
  \end{cases}
\end{flalign}
where $r$ indicates the distance from the center of the image, and $r_{\text{th}}$ and $g$ denote the hyper-parameters that define the focused area and depth gain, respectively.
We visualize this prior in Figure~\ref{fig:center_focus_prior}(b).
As shown in this figure, the prior facilitates the center area focus while promoting the surrounding area to be behind the focal plane.
We apply this prior to the generated depth $D^g$ as follows:
\begin{flalign}
  \label{eqn:prior}
  {\cal L}_p = \lambda_p \| D^g - D_p \|_2^2,
\end{flalign}
where $\lambda_p$ represents a weighting parameter.
In practice, we apply this only at the beginning of training to mitigate the negative effect triggered by the gap between $D^r$ and $D_p$.

\section{Experiments}
\label{sec:experiments}

\subsection{Experimental settings}
\label{subsec:experimental_settings}

We conducted four experiments to verify the effectiveness of AR-GANs from multiple perspectives:
a comparative study on unsupervised 3D representation learning (Section~\ref{subsec:comparative_study}),
ablation studies on DoF mixture learning and center focus prior (Section~\ref{subsec:ablation_study}),
portability analysis (Section~\ref{subsec:portability_analysis}), and
application in shallow DoF rendering (Section~\ref{subsec:application}).
Here, we explain the common settings and present the details of each in the following sections.

\smallskip\noindent\textbf{Datasets.}
We evaluated AR-GANs on three natural image datasets that cover various objects:
Oxford Flowers~\cite{MENilsbackICVGIP2008} (8189 flower images with 102 categories),
CUB-200-2011~\cite{CWahCUB2002011} (11788 bird images with 200 categories),
and FFHQ~\cite{TKarrasCVPR2019} (70000 face images).
To efficiently examine various cases, we resized the images to $64 \times 64$.
We also experimented on $128 \times 128$ images in some cases to confirm the dependency on image resolution (e.g., Figure~\ref{fig:concept}).

\smallskip\noindent\textbf{Metrics.}
To evaluate the visual fidelity of the generated images, we adopted the \textit{kernel inception distance (KID)}~\cite{MBinkowskiICLR2018},\footnote{We used KID because it has an unbiased estimator and complements the flaws of other representative metrics (i.e., Fr\'{e}chet inception distance (FID)~\cite{MHeuselNIPS2017} and inception score (IS)~\cite{MHeuselNIPS2017}).} which computes the maximum mean discrepancy between real and generated images within the Inception model~\cite{CSzegedyCVPR2016}.
When calculating scores, we generated 20000 images from each model.
Measuring depth and DoF accuracy directly is non-trivial because we aim to learn an unconditional model from unpaired and unlabeled natural images, and cannot obtain the ground truth.
Alternatively, we evaluated the depth accuracy by
(1) learning the depth estimator using pairs of images and depths generated by GANs,
(2) predicting the depths of real images using the learned depth estimator, and
(3) comparing the obtained results with the depths predicted by a state-of-the-art monocular depth estimator~\cite{KXianCVPR2020}, which is trained using stereo pairs in an external dataset.\footnote{We used the pretrained model provided by the authors: \url{https://github.com/KexianHust/Structure-Guided-Ranking-Loss}.}
We used \textit{scale-invariant depth error (SIDE)}~\cite{DEigenNIPS2014} to measure the difference.
In both metrics, the performance increased as the score decreased.
In all the experiments, we report the mean score with the standard deviation over three training runs.

\smallskip\noindent\textbf{Implementation.}
We implemented the model based on HoloGAN~\cite{TNguyenICCV2019}.
The generator has a StyleGAN-like architecture~\cite{TKarrasCVPR2019}.
In AR-GANs, 3D convolution used in HoloGAN is not required; hence, we replaced it with 2D convolution.
The discriminator has instance~\cite{DUlyanovArXiv2016} and spectral~\cite{TMiyatoICLR2018} normalizations.
The networks were trained using the Adam optimizer~\cite{DPKingmaICLR2015} with a non-saturating GAN loss~\cite{IGoodfellowNIPS2014}.

\begin{figure}[t]
  \centering
  \includegraphics[width=0.99\columnwidth]{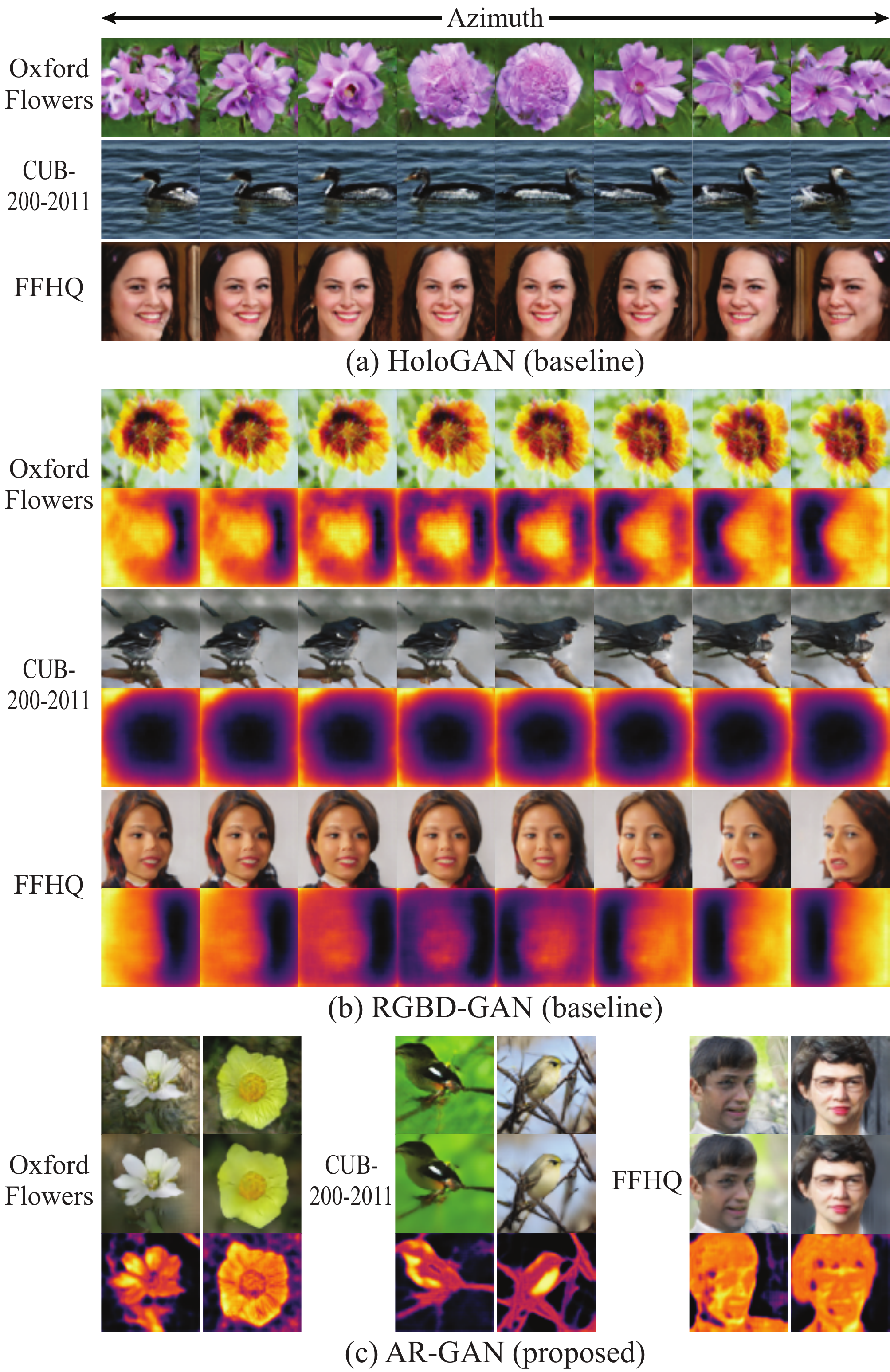}
  \caption{\textbf{Qualitative comparison among HoloGAN, RGBD-GAN, and AR-GAN.}
    HoloGAN, RGBD-GAN, and AR-GAN generate images, image and depth pairs, and tuples of deep and shallow DoF images and depths, respectively.}
  \label{fig:comparison_gans}
  \vspace{-4mm}
\end{figure}

\subsection{Comparative study}
\label{subsec:comparative_study}

First, we conducted a comparative study to clarify the difference between AR-GAN and previous fully unsupervised 3D representation learning.

\smallskip\noindent\textbf{Comparison models.}
We compared AR-GAN with \textit{HoloGAN}~\cite{TNguyenICCV2019} and \textit{RGBD-GAN}~\cite{ANoguchiICLR2020}, which are representative models in this category, as well as standard \textit{GAN}~\cite{IGoodfellowNIPS2014}.
As discussed in Section~\ref{sec:related_work}, HoloGAN/RGBD-GAN learns 3D representations using \textit{viewpoint cues}, whereas AR-GAN achieves this with \textit{focus cues}.
Hence, the applicable datasets are different, which we verified by using the three datasets.

\smallskip\noindent\textbf{Results.}
Examples of the generated images are presented in Figure~\ref{fig:comparison_gans}.
Here, the obtainable 3D representations and applicable datasets differ among the GANs.
Although HoloGAN and RGBD-GAN succeed in learning viewpoint-aware representations in FFHQ, they fail to do so in Oxford Flowers and CUB-200-2011, where viewpoint distributions are biased and viewpoint cues do not exist sufficiently.
By contrast, AR-GAN succeeds in learning the depth and DoF effect in all datasets because it can employ focus cues, which are present across all datasets.

\begin{table}
  \newcommand{\spm}[1]{{\tiny$\pm#1$}}
  \setlength{\tabcolsep}{0pt}
  \centering
  \scriptsize
  \begin{tabularx}{\columnwidth}{cCCC}
    \toprule
    KID$\times 10^3$$\downarrow$ & Oxford Flowers & CUB-200-2011 & FFHQ
    \\ \midrule
    GAN      & 11.71 \spm{0.68} & 15.04 \spm{0.14} & 6.97 \spm{0.30}
    \\ \midrule
    HoloGAN  & 11.30 \spm{0.37} & 14.68 \spm{0.51} & 6.89 \spm{0.38}
    \\
    RGBD-GAN & 12.04 \spm{0.35} & 14.92 \spm{0.49} & 6.73 \spm{0.26}
    \\ \midrule
    AR-GAN   & \textbf{11.23} \spm{0.36} & \textbf{14.30} \spm{0.56} & \textbf{5.75} \spm{0.19}
    \\ \bottomrule
  \end{tabularx}
  \setlength{\belowcaptionskip}{-5pt}
  \caption{\textbf{Comparison of KID$\times 10^3$$\downarrow$ among different GANs.}}
  \label{tab:comparison_kid_gans}
  \vspace{-2.5mm}
\end{table}

\begin{table}
  \newcommand{\spm}[1]{{\tiny$\pm#1$}}
  \setlength{\tabcolsep}{0pt}
  \centering
  \scriptsize
  \begin{tabularx}{\columnwidth}{cCCC}
    \toprule
    SIDE$\times 10^2$$\downarrow$ & Oxford Flowers & CUB-200-2011 & FFHQ
    \\ \midrule
    RGBD-GAN & 7.01 \spm{0.81} & 7.06 \spm{0.02} & 5.81 \spm{0.40}
    \\ \midrule
    AR-GAN   & \textbf{4.46} \spm{0.03} & \textbf{3.58} \spm{0.04} & \textbf{4.21} \spm{0.15}
    \\ \bottomrule
  \end{tabularx}
  \setlength{\belowcaptionskip}{-5pt}
  \caption{\textbf{Comparison of SIDE$\times 10^2$$\downarrow$ between RGBD-GAN and AR-GAN.}
    GAN and HoloGAN are not listed here because they cannot generate depth along with an image.}
  \label{tab:comparison_side_gans}
  \vspace{-3.5mm}
\end{table}

\begin{figure}
  \centering
  \includegraphics[width=0.99\columnwidth]{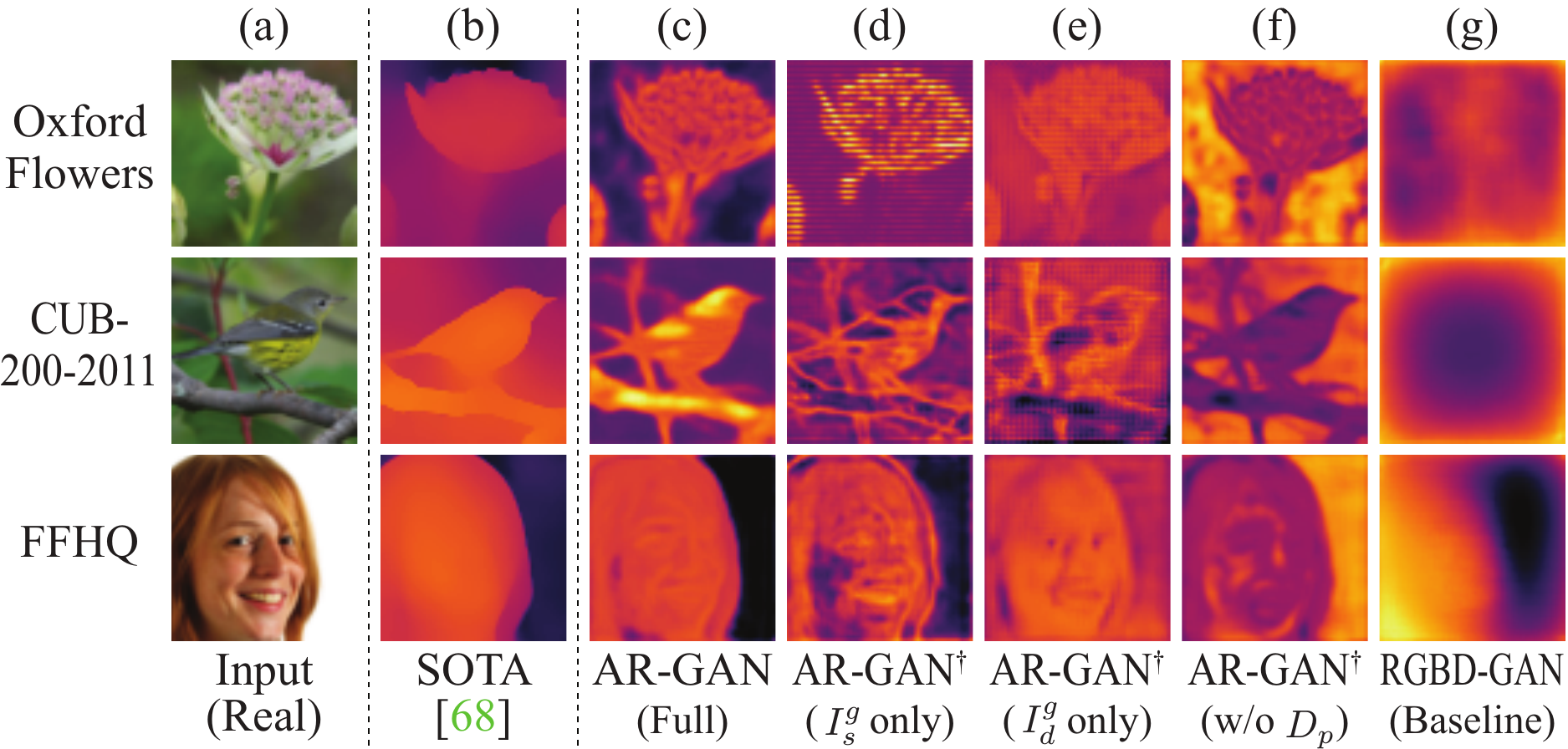}
  \caption{\textbf{Examples of predicted depths.}
    $^{\dag}$ indicates the ablated model.
    (c--g) Results obtained in a fully unsupervised setting.}
  \label{fig:depth_prediction}
  \vspace{-5mm}
\end{figure}

The KID comparison is summarized in Table~\ref{tab:comparison_kid_gans}.
We found that AR-GAN achieved comparable performance and did not incur a negative effect across all datasets.

The SIDE comparison is presented in Table~\ref{tab:comparison_side_gans}.
We found that AR-GAN outperforms RGBD-GAN in all datasets.
Examples of the predicted depths are presented in Figure~\ref{fig:depth_prediction}.
Although the predicted depths exhibit a lower resolution than those predicted by the supervised methods (e.g,~\cite{KXianCVPR2020}),\footnote{Note that $64 \times 64$ is the standard resolution for fully unsupervised learning methods (e.g., HoloGAN and RGBD-GAN), and applications to images with complex objects/backgrounds are challenging for them.} we found that AR-GAN (c) improves the details (e.g., flower details and tree branches) that disappear in~\cite{KXianCVPR2020} (b) and RGBD-GAN (g), thus benefiting from focus cues.

\subsection{Ablation study}
\label{subsec:ablation_study}

\begin{table}
  \newcommand{\spm}[1]{{\tiny$\pm#1$}}
  \setlength{\tabcolsep}{2pt}
  \centering
  \scriptsize
  \begin{tabularx}{\columnwidth}{clCCCC}
    \toprule
    \multicolumn{2}{c}{Oxford Flowers} & KID$\times 10^3$$\downarrow$ & SIDE$\times 10^2$$\downarrow$ & LPIPS$\downarrow$ & DSD$\uparrow$
    \\ \midrule
    $I_s^g$ only & $p_s = 1$ & 12.36 \spm{0.59} & 5.48 \spm{0.20} & 0.229 \spm{0.027} & 0.157 \spm{0.063}
    \\
    Mixture & $p_s = 0.75$ & 10.97 \spm{0.26} & 4.81 \spm{0.06} & 0.023 \spm{0.001} & 0.657 \spm{0.006}
    \\
    Mixture & $p_s = 0.5$ & 10.69 \spm{0.48} & 4.65 \spm{0.05} & 0.022 \spm{0.000} & 0.771 \spm{0.022}
    \\
    Mixture & $p_s = 0.25$ & 11.23 \spm{0.36} & 4.46 \spm{0.03} & 0.028 \spm{0.001} & 1.007 \spm{0.025}
    \\
    $I_d^g$ only & $p_s = 0$ & 11.58 \spm{0.37} & 4.56 \spm{0.20} & 0.113 \spm{0.013} & 0.446 \spm{0.065}
    \\ \midrule
    \multicolumn{2}{c}{L1} & 11.66 \spm{0.72} & 5.26 \spm{0.65} & 0.033 \spm{0.001} & 0.387 \spm{0.116}
    \\
    \multicolumn{2}{c}{Double discriminators} & 9.74 \spm{0.31} & 6.79 \spm{2.21} & 0.000 \spm{0.001} & 0.032 \spm{0.046}
    \\ \bottomrule
    \toprule
    \multicolumn{2}{c}{CUB-200-2011} & KID$\times 10^3$$\downarrow$ & SIDE$\times 10^2$$\downarrow$ & LPIPS$\downarrow$ & DSD$\uparrow$
    \\ \midrule
    $I_s^g$ only & $p_s = 1$ & 13.62 \spm{0.53} & 4.63 \spm{0.50} & 0.125 \spm{0.037} & 0.354 \spm{0.021}
    \\
    Mixture & $p_s = 0.75$ & 12.68 \spm{0.61} & 3.75 \spm{0.08} & 0.037 \spm{0.003} & 0.748 \spm{0.072}
    \\
    Mixture & $p_s = 0.5$ & 13.14 \spm{0.03} & 3.55 \spm{0.02} & 0.043 \spm{0.003} & 0.959 \spm{0.075}
    \\
    Mixture & $p_s = 0.25$ & 14.30 \spm{0.56} & 3.58 \spm{0.04} & 0.059 \spm{0.002} & 1.175 \spm{0.017}
    \\
    $I_d^g$ only & $p_s = 0$ & 14.58 \spm{0.56} & 5.94 \spm{0.70} & 0.115 \spm{0.019} & 0.193 \spm{0.012}
    \\ \midrule
    \multicolumn{2}{c}{L1} & 12.54 \spm{0.32} & 5.75 \spm{1.26} & 0.042 \spm{0.001} & 0.725 \spm{0.195}
    \\
    \multicolumn{2}{c}{Double discriminators} & 12.50 \spm{0.12} & 4.33 \spm{0.34} & 0.000 \spm{0.000} & 0.001 \spm{0.000}
    \\ \bottomrule
    \toprule
    \multicolumn{2}{c}{FFHQ} & KID$\times 10^3$$\downarrow$ & SIDE$\times 10^2$$\downarrow$ & LPIPS$\downarrow$ & DSD$\uparrow$
    \\ \midrule
    $I_s^g$ only & $p_s = 1$ & 5.75 \spm{0.44} & 6.00 \spm{0.35} & 0.097 \spm{0.011} & 0.296 \spm{0.018}
    \\
    Mixture & $p_s = 0.75$ & 5.67 \spm{0.23} & 4.38 \spm{0.10} & 0.009 \spm{0.001} & 0.757 \spm{0.177}
    \\
    Mixture & $p_s = 0.5$ & 5.75 \spm{0.19} & 4.21 \spm{0.15} & 0.009 \spm{0.001} & 0.769 \spm{0.119}
    \\
    Mixture & $p_s = 0.25$ & 6.17 \spm{0.08} & 4.68 \spm{0.33} & 0.010 \spm{0.001} & 0.583 \spm{0.071}
    \\
    $I_d^g$ only & $p_s = 0$ & 6.85 \spm{0.13} & 4.77 \spm{0.13} & 0.028 \spm{0.006} & 0.202 \spm{0.003}
    \\ \midrule
    \multicolumn{2}{c}{L1} & 5.82 \spm{0.21} & 4.82 \spm{0.09} & 0.015 \spm{0.004} & 0.466 \spm{0.045}
    \\
    \multicolumn{2}{c}{Double discriminators} & 6.20 \spm{0.08} & 5.20 \spm{0.47} & 0.000 \spm{0.000} & 0.000 \spm{0.000}
    \\ \bottomrule
  \end{tabularx}
  \setlength{\belowcaptionskip}{-5pt}
  \caption{\textbf{Comparison of KID$\times 10^3$$\downarrow$, SIDE$\times 10^2$$\downarrow$, LPIPS$\downarrow$, and DSD$\uparrow$ among AR-GANs with different learning methods.}}
  \label{tab:ablation_dof_mixture_learning}
  \vspace{-5mm}
\end{table}

\subsubsection{Ablation study on DoF mixture learning}
\label{subsubsec:ablation_study_dof_mixture_learning}

\smallskip\noindent\textbf{Metrics.}
We first evaluated the importance of the DoF mixture learning.
As mentioned in Section~\ref{subsec:experimental_settings}, measuring depth and DoF accuracy directly is non-trivial; therefore, we further adopted two metrics along with KID and SIDE:
\textit{(1) Learned perceptual image patch similarity (LPIPS)}~\cite{RZhangCVPR2018} computes the distance between two images in the CNN feature space and is demonstrated to exhibit a high correlation with human perceptual similarity~\cite{RZhangCVPR2018}.
We adopted LPIPS to measure the perceptual similarity between pairs of $I_d^g$ and $I_s^g$.
LPIPS is expected to be moderately small because the content is preserved before and after the application of aperture rendering.
\textit{(2) Depth standard deviation (DSD)} is the standard deviation of the generated depths.
Our objective is to learn a meaningful depth that can yield a plausible DoF effect.
When depth learning is successful, DSD is expected to be sufficiently large.

\smallskip\noindent\textbf{Comparison models.}
We conducted the analysis from two perspectives.
(1) We evaluated the effect of the $p_s$ value, which indicates the rate of using shallow DoF images in the DoF mixture learning (Equation~\ref{eqn:ar-gan}).
(2) We tested two possible alternatives:
\textit{L1}, which uses L1 loss to guide $I_d^g$ closer to $I_s^g$,
and \textit{double discriminators}, which adopts two discriminators, for $I_d^g$ and $I_s^g$, respectively.
This facilitates both $p^g(I_d)$ and $p^g(I_s)$ to coincide with the overall real distribution $p^r(I)$.

\smallskip\noindent\textbf{Results.}
A comparison of the scores is summarized in Table~\ref{tab:ablation_dof_mixture_learning}.
Our main findings are two-fold:

\smallskip\textit{(1) Effect of value of $p_s$.}
We found that some fluctuations exist in the KID; however, in all cases, the scores are comparable to those of the other GANs presented in Table~\ref{tab:comparison_kid_gans}.
This indicates that AR-GANs can generate plausible images regardless of $p_s$.
By contrast, SIDE, LPIPS, and DSD are affected by $p_s$.
SIDE tends to improve when the DoF mixture learning is adopted.\footnote{The sole ``$I_d^g$ only'' case in Oxford Flowers is an exception.
  In this case, $D^g$ is not regularized by aperture rendering; however, weight sharing between $G_I$ and $G_D$ (Section~\ref{subsec:overall_pipeline}) aids the depth learning.
  This strategy exhibits dataset dependency and fails in the other datasets.}
This is because in the DoF mixture learning, we can encourage $I_d^g$ and $I_s^g$ to approximate $I_d^r$ and $I_s^r$, respectively, as well as facilitate $D^g$ learning, which is the source that connects them.
Examples of predicted depths (Figure~\ref{fig:depth_prediction} (c--e)) also validate the effectiveness of DoF mixture learning.
Regarding LPIPS and DSD, as LPIPS increases, and DSD successively decreases when $p_s = 1$ or $p_s = 0$.
This indicates that the DoF mixture learning is required to manage LPIPS and DSD.
Among AR-GANs with DoF mixture learning (i.e., $p_s \in \{ 0.25, 0.5, 0.75\}$), there is a trade-off and dataset dependency relative to LPIPS and DSD.
Consider the score balance, we set $p_s = 0.25$ for Oxford Flowers and CUB-200-2011 and $p_s = 0.5$ for FFHQ in other experiments.

\smallskip\textit{(2) Comparison with alternatives}.
Although L1 achieves reasonable LPIPS, it deteriorates SIDE and DSD more than those in the DoF mixture learning.
This result indicates that this method disrupts depth learning.
Double discriminators facilitate both $p^g(I_d)$ and $p^g(I_s)$ to coincide with $p^r(I)$.
Consequently, LPIPS and DSD approach zero, and SIDE depreciates.
This result verifies the importance of mixing $I_d^g$ and $I_s^g$ when learning $p^r(I)$.

\begin{figure}[t]
  \centering
  \includegraphics[width=0.95\columnwidth]{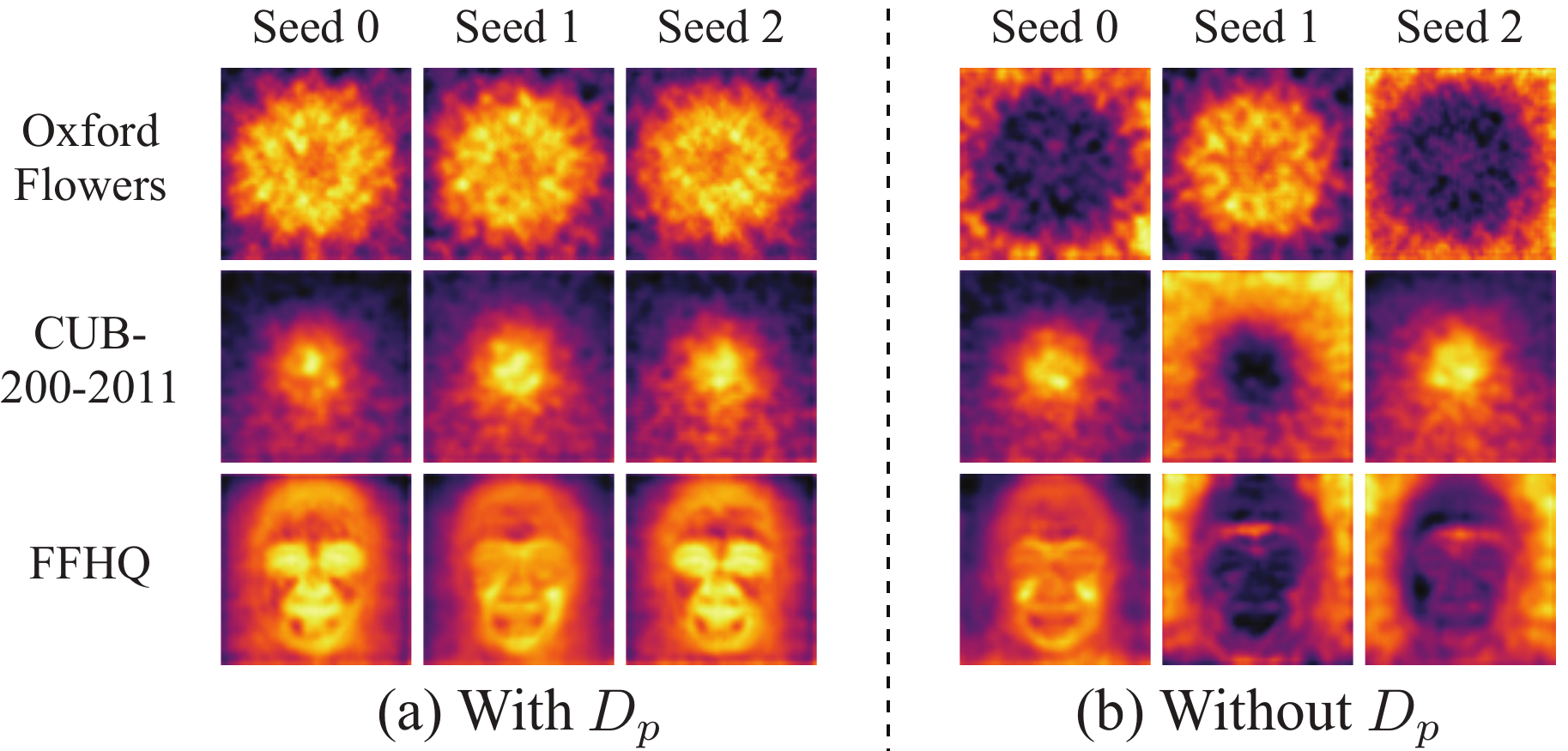}
  \vspace{-1mm}
  \caption{\textbf{Comparison of AD with and without $D_p$.}}
  \label{fig:comparison_ad}
  \vspace{-1mm}
\end{figure}

\begin{table}
  \newcommand{\spm}[1]{{\tiny$\pm#1$}}
  \setlength{\tabcolsep}{2pt}
  \centering
  \scriptsize
  \begin{tabularx}{\columnwidth}{cCCCC}
    \toprule
    Oxford Flowers & KID$\times 10^3$$\downarrow$ & SIDE$\times 10^2$$\downarrow$ & LPIPS$\downarrow$ & DSD$\uparrow$
    \\ \midrule
    W/ $D_p$  & 11.23 \spm{0.36} & 4.46 \spm{0.03} & 0.028 \spm{0.001} & 1.007 \spm{0.025}
    \\
    W/o $D_p$ & 10.69 \spm{0.24} & 6.78 \spm{1.58} & 0.026 \spm{0.002} & 0.915 \spm{0.137}
    \\ \bottomrule
    \toprule
    CUB-200-2011 & KID$\times 10^3$$\downarrow$ & SIDE$\times 10^2$$\downarrow$ & LPIPS$\downarrow$ & DSD$\uparrow$
    \\ \midrule
    W/ $D_p$  & 14.30 \spm{0.56} & 3.58 \spm{0.04} & 0.059 \spm{0.002} & 1.175 \spm{0.017}
    \\
    W/o $D_p$ & 13.96 \spm{0.63} & 4.86 \spm{1.84} & 0.062 \spm{0.004} & 1.183 \spm{0.059}
    \\ \bottomrule
    \toprule
    FFHQ & KID$\times 10^3$$\downarrow$ & SIDE$\times 10^2$$\downarrow$ & LPIPS$\downarrow$ & DSD$\uparrow$
    \\ \midrule
    W/ $D_p$  & 5.75 \spm{0.19} & 4.21 \spm{0.15} & 0.009 \spm{0.001} & 0.769 \spm{0.119}
    \\
    W/o $D_p$ & 5.72 \spm{0.10} & 6.70 \spm{1.88} & 0.009 \spm{0.001} & 0.851 \spm{0.057}
    \\ \bottomrule
  \end{tabularx}
  \setlength{\belowcaptionskip}{-5pt}
  \caption{\textbf{Comparison of KID$\times 10^3$$\downarrow$, SIDE$\times 10^2$$\downarrow$, LPIPS$\downarrow$, and DSD$\uparrow$ among AR-GANs with and without $D_p$.}}
  \label{tab:ablation_center_focus_prior}
  \vspace{-5mm}
\end{table}

\subsubsection{Ablation study on center focus prior}
\label{subsubsec:ablation_study_center_focus_prior}

\smallskip\noindent\textbf{Metrics.}
We evaluated the necessity of the center focus prior $D_p$.
To assess the overall tendency of each pixel to represent the foreground or background blur, we calculated the \textit{average depth (AD)}, i.e., the pixel-wise average of the generated depths.
To validate the learning consistency, we compared the results over three training runs.

\smallskip\noindent\textbf{Results.}
The results are presented in Figure~\ref{fig:comparison_ad}.
We found that we can obtain constant results across training runs when $D_p$ is adopted.
In all results, the center is focused, while the surroundings are behind the focal plane.
By contrast, when we eliminate $D_p$, the foreground and background are turned over, depending on the initialization.
These results indicate that $D_p$ is beneficial in determining the learning direction.

The comparison of the scores is summarized in Table~\ref{tab:ablation_center_focus_prior}.
We found that KID, LPIPS, and DSD are comparable across all datasets.
We deduce that $D_p$ is solely adopted at the beginning of training; therefore, it does not disrupt the entire training.
By contrast, SIDE depreciates when $D_p$ is not implemented.
This occurs because the foreground and background are reversed, as shown in Figure~\ref{fig:depth_prediction}(f).

\subsection{Portability analysis}
\label{subsec:portability_analysis}

\begin{figure}[t]
  \centering
  \includegraphics[width=\columnwidth]{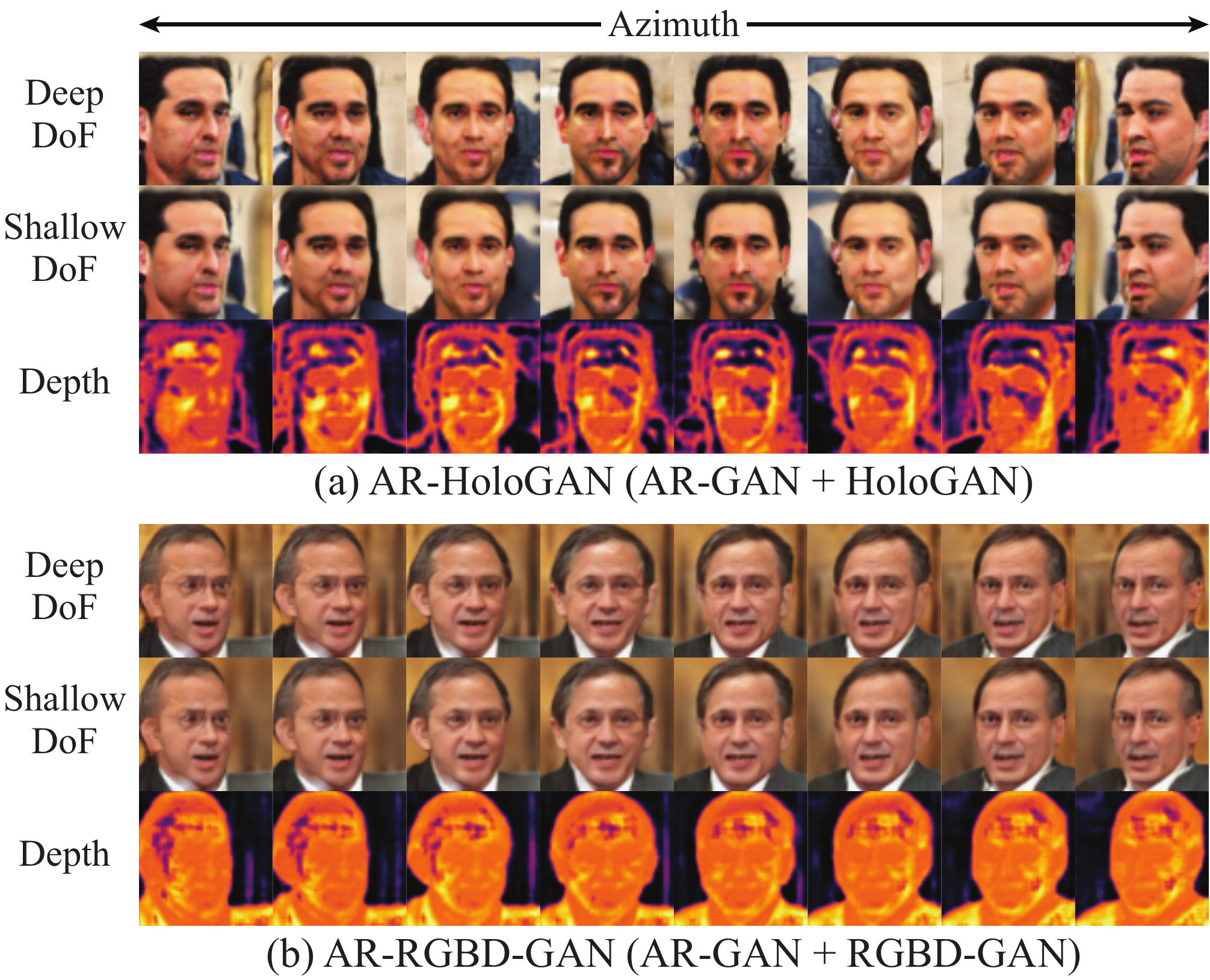}
  \caption{\textbf{Examples of data generated using AR-HoloGAN and AR-RGBD-GAN.}
    The viewpoint change in the horizontal direction is obtained by the HoloGAN/RGBD-GAN function, whereas the DoF change and depth in the vertical direction are obtained by the AR-GAN function.}
  \label{fig:combination}
  \vspace{-2mm}
\end{figure}

As presented in Section~\ref{subsec:comparative_study}, the obtainable representations differ between HoloGAN/RGBD-GAN and AR-GAN.
An interesting approach would be learning these representations jointly by combining them.
A significant property of AR-GAN is \textit{portability}, i.e., it is easy to incorporate into other GANs.
Specifically, we can achieve this simply by adding aperture rendering on top of HoloGAN/RGBD-GAN and training it with the DoF mixture training and a center focus prior.
One requirement is that a dataset should satisfy the assumptions of both models; i.e., a dataset should include diverse viewpoint images along with various DoF images.
Among the datasets described above, only FFHQ satisfies this requirement.
Hence, we solely evaluated \textit{AR-HoloGAN} (AR-GAN + HoloGAN) and \textit{AR-RGBD-GAN} (AR-GAN + RGBD-GAN) on this dataset.

\smallskip\noindent\textbf{Results.}
Examples of generated data are shown in Figure~\ref{fig:combination}.
As shown in this figure, we can jointly control both viewpoints and the DoF effect with the HoloGAN/RGBD-GAN and AR-GAN functions.
As a reference, we also calculated the KID scores.
The scores for AR-HoloGAN and AR-RGBD-GAN were $5.70 \pm 0.32$ and $5.43 \pm 0.22$, respectively.
These are better than the scores of the original AR-GAN, HoloGAN, and RGBD-GAN (Table~\ref{tab:comparison_kid_gans}).

\begin{figure}[t]
  \centering
  \includegraphics[width=\columnwidth]{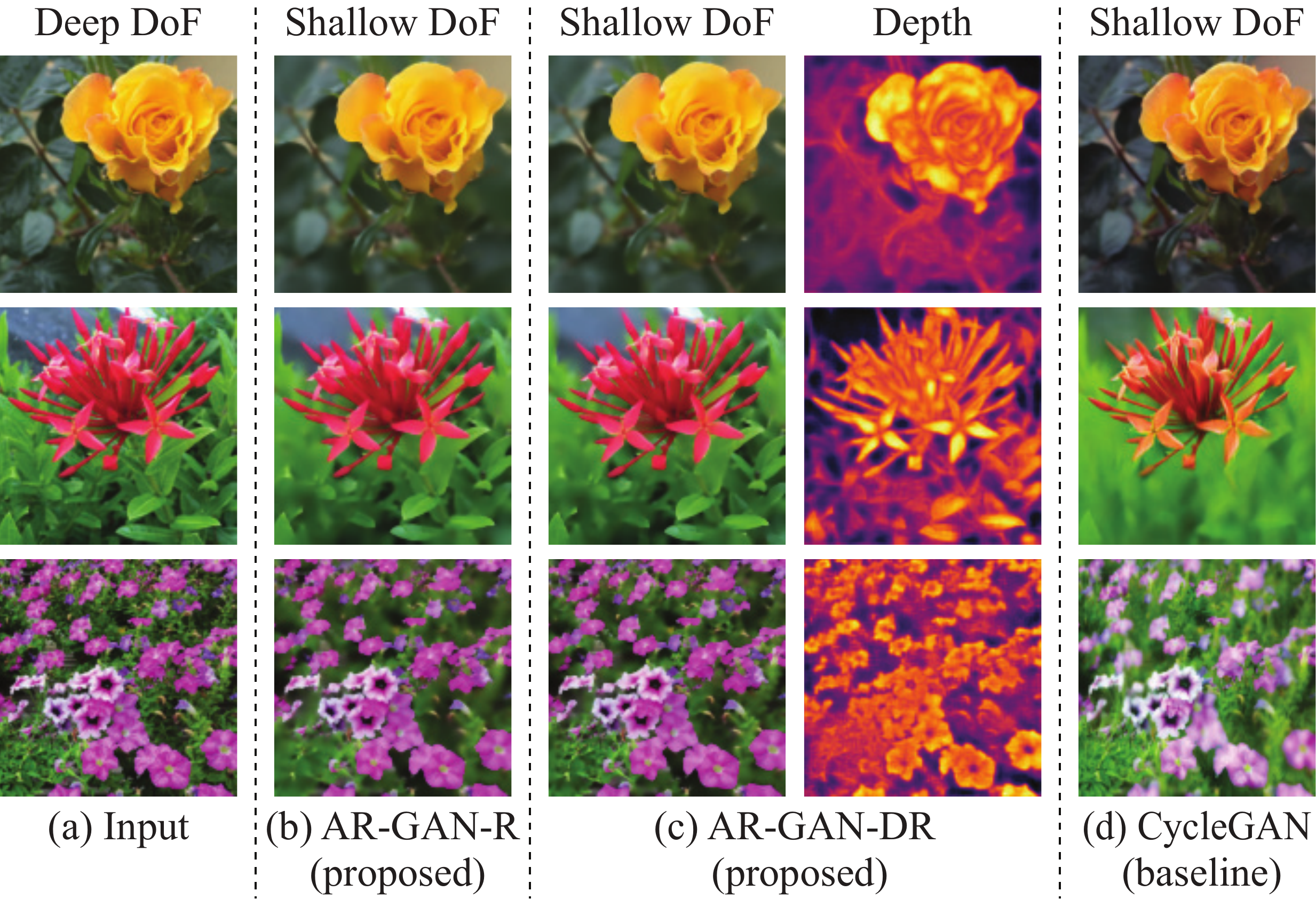}
  \caption{\textbf{Examples of shallow DoF rendering.}}
  \label{fig:shallow_dof_rendering}
  \vspace{-3mm}
\end{figure}

\subsection{Application in shallow DoF rendering}
\label{subsec:application}

Finally, we demonstrate the applicability of AR-GAN in shallow DoF rendering.
After training, AR-GAN can synthesize tuples of $(I_d^g, D^g, I_s^g)$ from random noise.
By utilizing this, we learn a shallow DoF renderer $I_d \rightarrow I_s$ using pairs of $(I_d^g, I_s^g)$.
We call this approach \textit{AR-GAN-R}.
As another approach, we learn a depth estimator $I_d \rightarrow D$ using pairs of $(I_d^g, D_g)$.
By employing the learned depth estimator, we estimate $D$ from $I_d$ and then render $I_s$ from $(I_d, D)$ using $R$ in AR-GAN.
We call this approach \textit{AR-GAN-DR}.

\smallskip\noindent\textbf{Dataset.}
We used Oxford Flowers and AR-GAN-generated images to train AR-GAN and AR-GAN-R/DR, respectively.
To confirm generality, we conducted a test on a different dataset, including flower photos taken by smartphones, which were used in the CycleGAN study~\cite{JYZhuICCV2017}.

\smallskip\noindent\textbf{Comparison model.}
To the best of our knowledge, no previous method can learn the DoF effect from natural images in the same setting as ours (i.e., \textit{without} additional supervision and a predefined model).
Therefore, as a baseline, we used CycleGAN~\cite{JYZhuICCV2017}, which trains a shallow DoF renderer $I_d \rightarrow I_s$ using \textit{set-level} supervision (i.e., supervision of whether each image is a deep or shallow DoF image).\footnote{We used the pretrained model provided by the authors: \url{https://github.com/junyanz/pytorch-CycleGAN-and-pix2pix}.}

\smallskip\noindent\textbf{Results.}
Examples of the rendered images are presented in Figure~\ref{fig:shallow_dof_rendering}.
We found that CycleGAN often yields unnecessary changes (e.g., color change), whereas AR-GAN-R/DR does not.
We infer that the aperture rendering mechanism in AR-GAN contributes to this phenomenon.
In addition, AR-GAN-DR can estimate the depth simultaneously.

\section{Conclusion}
\label{sec:conclusion}

We proposed a novel family of GANs, AR-GANs, which can learn depth and DoF effect from unconstrained natural images.
To achieve this, we incorporated aperture rendering into GANs and developed DoF mixture learning and a center focus prior to address the ambiguities triggered by the unsupervised setting.
Via comparative and ablation studies, we elucidated the differences from previous GANs and the significance of the proposed techniques.
We demonstrated that AR-GANs are compatible and complementary to previous GANs by combining AR-GANs with HoloGAN/RGBD-GAN.
Finally, we demonstrated the applicability of AR-GANs in shallow DoF rendering.
Despite their applications in photos, several deep generative models do not utilize focus cues.
In the future, we expect that our findings will facilitate further studies on such models.

\clearpage
\renewcommand{\baselinestretch}{0.925}
{\small
\bibliographystyle{ieee_fullname}
\bibliography{refs}
}

\clearpage
\renewcommand{\baselinestretch}{1}
\appendix
\section*{Appendix}

In this appendix, we provide further analyses (Appendix~\ref{sec:further_analyses}), extended results (Appendix~\ref{sec:extended_results}), and implementation details (Appendix~\ref{sec:implementation_details}) regarding our study.

\section{Further analyses}
\label{sec:further_analyses}

In this section, we provide three additional analyses for a deeper understanding of metrics and the proposed model.
\begin{itemize}
\item Appendix~\ref{subsec:further_analysis_of_metrics}:
  Further analysis of metrics.
\item Appendix~\ref{subsec:further_analysis_dof_mixture_learning}:
  Further analysis of DoF mixture learning.
  This is a detailed analysis of the experiments presented in Section~\ref{subsubsec:ablation_study_dof_mixture_learning}.
\item Appendix~\ref{subsec:further_analysis_center_focus_prior}:
  Further analysis of the center focus prior.
  This is a detailed analysis of the experiments presented in Section~\ref{subsubsec:ablation_study_center_focus_prior}.
\end{itemize}

\begin{table*}
  \newcommand{\spm}[1]{{\tiny$\pm#1$}}
  \setlength{\tabcolsep}{3pt}
  \centering
  \scriptsize
  \begin{subtable}{\textwidth}
    \caption{Oxford Flowers}
    \vspace{-2mm}
    \begin{tabularx}{\textwidth}{cclCCCCCCC}
      \toprule
      No. & \multicolumn{2}{c}{Learning method} & KID$\times 10^3$$\downarrow$ & KID$_{I_s^g}$$\times 10^3$$\downarrow$ & KID$_{I_d^g}$$\times 10^3$$\downarrow$ & SIDE$\times 10^2$$\downarrow$ & LPIPS$\downarrow$ & SSIM$\uparrow$ & DSD$\uparrow$
      \\ \midrule
      1 & $I_s^g$ only & $p_s = 1$
      & 12.36 \spm{0.59}
      & 12.46 \spm{0.58}
      & 293.56 \spm{44.92}
      & 5.48 \spm{0.20}
      & 0.229 \spm{0.027}
      & 0.406 \spm{0.049}
      & 0.157 \spm{0.063}
      \\
      2 & Mixture & $p_s = 0.75$
      & 10.97 \spm{0.26}
      & 13.74 \spm{0.47}
      & 26.50 \spm{1.59}
      & 4.81 \spm{0.06}
      & 0.023 \spm{0.001}
      & 0.943 \spm{0.002}
      & 0.657 \spm{0.006}
      \\
      3 & Mixture & $p_s = 0.5$
      & 10.69 \spm{0.48}
      & 18.98 \spm{1.24}
      & 17.58 \spm{0.53}
      & 4.65 \spm{0.05}
      & 0.022 \spm{0.000}
      & 0.940 \spm{0.002}
      & 0.771 \spm{0.022}
      \\
      4 & Mixture & $p_s = 0.25$
      & 11.23 \spm{0.36}
      & 30.61 \spm{2.99}
      & 12.91 \spm{0.38}
      & 4.46 \spm{0.03}
      & 0.028 \spm{0.001}
      & 0.919 \spm{0.002}
      & 1.007 \spm{0.025}
      \\
      5 & $I_d^g$ only & $p_s = 0$
      & 11.58 \spm{0.37}
      & 36.63 \spm{5.11}
      & 11.51 \spm{0.41}
      & 4.56 \spm{0.20}
      & 0.113 \spm{0.013}
      & 0.870 \spm{0.020}
      & 0.446 \spm{0.065}
      \\ \midrule
      6 & \multicolumn{2}{c}{L1}
      & 11.66 \spm{0.72}    
      & 11.81 \spm{0.74}
      & 47.59 \spm{2.54}
      & 5.26 \spm{0.65}
      & 0.033 \spm{0.001}
      & 0.932 \spm{0.002}
      & 0.387 \spm{0.116}
      \\
      7 & \multicolumn{2}{c}{Double discriminators}
      & 9.74 \spm{0.31}    
      & 9.79 \spm{0.38}
      & 9.88 \spm{0.30}
      & 6.79 \spm{2.21}
      & 0.000 \spm{0.001}
      & 0.999 \spm{0.001}
      & 0.032 \spm{0.046}
      \\ \bottomrule
    \end{tabularx}    
  \end{subtable}
  \begin{subtable}{\textwidth}
    \vspace{2mm}
    \caption{CUB-200-2011}
    \vspace{-2mm}
    \begin{tabularx}{\textwidth}{cclCCCCCCC}
      \toprule
      No. & \multicolumn{2}{c}{Learning method} & KID$\times 10^3$$\downarrow$ & KID$_{I_s^g}$$\times 10^3$$\downarrow$ & KID$_{I_d^g}$$\times 10^3$$\downarrow$ & SIDE$\times 10^2$$\downarrow$ & LPIPS$\downarrow$ & SSIM$\uparrow$ & DSD$\uparrow$
      \\ \midrule
      1 & $I_s^g$ only & $p_s = 1$
      & 13.62 \spm{0.53}
      & 13.71 \spm{0.58}
      & 72.63 \spm{25.71}
      & 4.63 \spm{0.50}
      & 0.125 \spm{0.037}
      & 0.786 \spm{0.029}
      & 0.354 \spm{0.021}
      \\
      2 & Mixture & $p_s = 0.75$
      & 12.68 \spm{0.61}
      & 12.30 \spm{0.45}
      & 22.64 \spm{0.37}
      & 3.75 \spm{0.08}
      & 0.037 \spm{0.003}
      & 0.921 \spm{0.006}
      & 0.748 \spm{0.072}
      \\
      3 & Mixture & $p_s = 0.5$
      & 13.14 \spm{0.03}
      & 13.20 \spm{0.43}
      & 18.15 \spm{0.07}
      & 3.55 \spm{0.02}
      & 0.043 \spm{0.003}
      & 0.909 \spm{0.007}
      & 0.959 \spm{0.075}
      \\
      4 & Mixture & $p_s = 0.25$
      & 14.30 \spm{0.56}
      & 18.28 \spm{0.73}
      & 16.19 \spm{0.63}
      & 3.58 \spm{0.04}
      & 0.059 \spm{0.002}
      & 0.877 \spm{0.003}
      & 1.175 \spm{0.017}
      \\
      5 & $I_d^g$ only & $p_s = 0$
      & 14.58 \spm{0.56}
      & 28.63 \spm{2.14}
      & 14.85 \spm{0.72}
      & 5.94 \spm{0.70}
      & 0.115 \spm{0.019}
      & 0.908 \spm{0.010}
      & 0.193 \spm{0.012}
      \\ \midrule
      6 & \multicolumn{2}{c}{L1}
      & 12.54 \spm{0.32}
      & 12.59 \spm{0.45}
      & 33.70 \spm{0.84}
      & 5.75 \spm{1.26}
      & 0.042 \spm{0.001}
      & 0.910 \spm{0.001}
      & 0.725 \spm{0.195}
      \\
      7 & \multicolumn{2}{c}{Double discriminators}
      & 12.50 \spm{0.12}
      & 12.56 \spm{0.20}
      & 12.61 \spm{0.29}
      & 4.33 \spm{0.34}
      & 0.000 \spm{0.000}
      & 1.000 \spm{0.000}
      & 0.001 \spm{0.000}
      \\ \bottomrule
    \end{tabularx}    
  \end{subtable}
  \begin{subtable}{\textwidth}
    \vspace{2mm}
    \caption{FFHQ}
    \vspace{-2mm}
    \begin{tabularx}{\textwidth}{cclCCCCCCC}
      \toprule
      No. & \multicolumn{2}{c}{Learning method} & KID$\times 10^3$$\downarrow$ & KID$_{I_s^g}$$\times 10^3$$\downarrow$ & KID$_{I_d^g}$$\times 10^3$$\downarrow$ & SIDE$\times 10^2$$\downarrow$ & LPIPS$\downarrow$ & SSIM$\uparrow$ & DSD$\uparrow$
      \\ \midrule
      1 & $I_s^g$ only & $p_s = 1$
      & 5.75 \spm{0.44}
      & 5.82 \spm{0.48}
      & 74.36 \spm{7.41}
      & 6.00 \spm{0.35}
      & 0.097 \spm{0.011}
      & 0.808 \spm{0.034}
      & 0.296 \spm{0.018}
      \\
      2 & Mixture & $p_s = 0.75$
      & 5.67 \spm{0.23}
      & 6.05 \spm{0.14}
      & 14.50 \spm{0.65}
      & 4.38 \spm{0.10}
      & 0.009 \spm{0.001}
      & 0.967 \spm{0.006}
      & 0.757 \spm{0.177}
      \\
      3 & Mixture & $p_s = 0.5$
      & 5.75 \spm{0.19}
      & 7.53 \spm{0.43}
      & 9.34 \spm{0.22}
      & 4.21 \spm{0.15}
      & 0.009 \spm{0.001}
      & 0.966 \spm{0.006}
      & 0.769 \spm{0.119}
      \\
      4 & Mixture & $p_s = 0.25$
      & 6.17 \spm{0.08}
      & 11.55 \spm{0.70}
      & 7.53 \spm{0.41}
      & 4.68 \spm{0.33}
      & 0.010 \spm{0.001}
      & 0.968 \spm{0.004}
      & 0.583 \spm{0.071}
      \\
      5 & $I_d^g$ only & $p_s = 0$
      & 6.85 \spm{0.13}
      & 10.16 \spm{0.29}
      & 6.94 \spm{0.10}
      & 4.77 \spm{0.13}
      & 0.028 \spm{0.006}
      & 0.976 \spm{0.003}
      & 0.202 \spm{0.003}
      \\ \midrule
      6 & \multicolumn{2}{c}{L1}
      & 5.82 \spm{0.21}
      & 5.83 \spm{0.23}
      & 24.89 \spm{2.70}
      & 4.82 \spm{0.09}
      & 0.015 \spm{0.004}
      & 0.967 \spm{0.005}
      & 0.466 \spm{0.045}
      \\
      7 & \multicolumn{2}{c}{Double discriminators}
      & 6.20 \spm{0.08}
      & 6.26 \spm{0.06}
      & 6.19 \spm{0.07}
      & 5.20 \spm{0.47}
      & 0.000 \spm{0.000}
      & 1.000 \spm{0.000}
      & 0.000 \spm{0.000}
      \\ \bottomrule
    \end{tabularx}
  \end{subtable}
  \vspace{0mm}
  \caption{\textbf{Comparison of KID$\times 10^3$$\downarrow$, KID$_{I_s^g}$$\times 10^3$$\downarrow$, KID$_{I_d^g}$$\times 10^3$$\downarrow$, SIDE$\times 10^2$$\downarrow$, LPIPS$\downarrow$, SSIM$\uparrow$, and DSD$\uparrow$ among AR-GANs with different learning methods.}
    This is an extended version of Table~\ref{tab:ablation_dof_mixture_learning}.}
  \label{tab:ablation_dof_mixture_learning_ex}
  \vspace{-2mm}
\end{table*}

\subsection{Further analysis of metrics}
\label{subsec:further_analysis_of_metrics}

As discussed in Section~\ref{subsec:experimental_settings}, because of our objective and formulation, that is, learning an unconditional model from unpaired and unlabeled natural images, it is nontrivial to prepare the ground truth or construct a metric that can measure depth and DoF accuracy directly.
As alternatives, in the ablation study (Section~\ref{subsec:ablation_study}), we adopted two metrics: \textit{learned perceptual image patch similarity (LPIPS)}~\cite{RZhangCVPR2018} and \textit{depth standard deviation (DSD)}, along with \textit{kernel inception distance (KID)}~\cite{MBinkowskiICLR2018} and \textit{scale-invariant depth error (SIDE)}~\cite{DEigenNIPS2014}.
For further clarification, in this subsection, we first evaluate the performance of three additional metrics along with the aforementioned metrics (Section~\ref{subsubsec:evaluation_on_additional_metrics}), and then discuss their relationships (Section~\ref{subsubsec:relationships_among_metrics}).

\subsubsection{Evaluation on additional metrics}
\label{subsubsec:evaluation_on_additional_metrics}

\smallskip\noindent\textbf{Additional metrics.}
We adopted three additional metrics.

\smallskip\textit{(1) Structural similarity index measure (SSIM)~\cite{ZWangTIP2004}.}
SSIM is a representative traditional (or non-DNN-based) metric for predicting perceived quality.
This metric measures the distance between two images; a larger value indicates higher similarity.
SSIM is a possible alternative to LPIPS, which was used in Section~\ref{subsec:ablation_study}.
The reason for using LPIPS is that a previous study~\cite{RZhangCVPR2018} demonstrated that LPIPS has a higher correlation with human perceptual judgments than SSIM.
We consider this characteristic to be useful for measuring the validity of the DoF effect where, between deep and shallow DoF images, the appearance of focused targets is nearly similar, but that of unfocused regions is relatively different.
To validate this consideration, we will compare LPIPS and SSIM later.

\smallskip\textit{(2) KID$_{I_s^g}$.}
We calculated the KID score between \textit{all real} images (i.e., $I^r \sim p^r(I)$) and \textit{generated shallow DoF} images (i.e., $I_s^g = R(G_I(z), G_D(z))$, where $z \sim p(z)$).
When $I^r$ includes both deep (i.e., $I_d^r \sim p^r(I_d)$) and shallow (i.e., $I_s^r \sim p^r(I_s)$) DoF images, it is expected that KID$_{I_s^g}$ is moderately small but larger than KID, which measures the distance between \textit{all real} (i.e., $I^r \sim p^r(I)$) images and \textit{all generated} images (i.e., $I^g \sim p^g(I)$).\footnote{When calculating KID for AR-GANs, we used the DoF mixture setting that was used in training.
  Therefore, in Table~\ref{tab:ablation_dof_mixture_learning}, the rates of $I_d^g$ and $I_s^g$ that were used to calculate the KID vary according to the model.}

\smallskip\textit{(3) KID$_{I_d^g}$.}
We computed the KID score between \textit{all real} images (i.e., $I^r \sim p^r(I)$) and \textit{generated deep DoF} images (i.e., $I_d^g = G_I(z)$, where $z \sim p(z)$).
This metric has a similar characteristic as KID$_{I_s^g}$.

\smallskip\noindent\textbf{Results.}
We applied the aforementioned metrics to the models presented in Table~\ref{tab:ablation_dof_mixture_learning}.
Table~\ref{tab:ablation_dof_mixture_learning_ex} summarizes the results.
Our main findings are three-fold:

\smallskip\textit{(1) Comparison of KID, KID$_{I_s^g}$, and KID$_{I_d^g}$ among AR-GANs with different $p_s$ (Nos.~1--5).}
We found that as $p_s$ (i.e., the probability that a shallow DoF image is rendered) decreases, KID$_{I_s^g}$ increases and KID$_{I_d^g}$ decreases.
In particular, when comparing KID$_{I_s^g}$ and KID$_{I_d^g}$, the KID for the dominant image (i.e., one used at a higher frequency in training) obtains a smaller value than the KID for the non-dominant image.
The KID for the non-dominant image is also moderately small when the image is used in training (i.e., $p_s \in \{ 0.75, 0.5, 0.25 \}$ (Nos.~2--4)).
By contrast, when either $I_s^g$ or $I_d^g$ is used for training (i.e., $p_s \in \{ 1, 0 \}$ (Nos.~1 and 5)), the KID for the unused one is relatively high.
These results imply that DoF mixture learning is useful for matching both $p^g(I_d)$ and $p^g(I_s)$ with parts of the real distribution, which are expected to be $p^r(I_d)$ and $p^r(I_s)$, respectively.

\smallskip\textit{(2) Comparison of KID, KID$_{I_s^g}$, and KID$_{I_d^g}$ between AR-GANs with different learning methods (Nos.~6 and 7).}
When an explicit distance metric (L1 in this experiment (No.~6)) is used, it is encouraged to compare generated deep DoF images (i.e., $I_d^g$) with generated shallow DoF images (i.e., $I_s^g$).
However, an adversarial loss is imposed only on $I_s^g$; consequently, its KID$_{I_d^g}$ is significantly larger than that of AR-GANs with DoF mixture learning (Nos.~2--4).
This suggests that, in this case, the visual fidelity of $I_d^g$ may not be high, even when the LPIPS is small.

When the double discriminators (No.~7) are used, both KID$_{I_s^g}$ and KID$_{I_d^g}$ approach KID.
This result indicates that both $p^g(I_s)$ and $p^g(I_d)$ are encouraged to coincide with the overall real distribution $p^r(I)$, which consists of $p^r(I_s)$ and $p^r(I_d)$.
This phenomenon can also be explained by the fact that LPIPS and DSD are almost 0, and SSIM is close to 1.

From these results, we conclude that DoF mixture learning is more reasonable than L1 and double discriminators when a dataset includes both deep and shallow DoF images (Assumption~\ref{assumption:ar-gan}), and we aim to selectively represent the real image distribution according to the DoF strength.

\smallskip\textit{(3) Comparison of LPIPS and SSIM.}
In most cases, as LPIPS decreases, SSIM increases.
We would like to note that the similarity is higher when the LPIPS value is small, as well as when the SSIM value is large.
However, in some cases (for example, between Nos.~4 and 5 in CUB-200-2011), this tendency does not hold.
We discuss this in Appendix~\ref{subsubsec:relationships_among_metrics}.

\subsubsection{Relationships among metrics}
\label{subsubsec:relationships_among_metrics}

In the results in Table~\ref{tab:ablation_dof_mixture_learning_ex}, there are two nontrivial relationships among metrics: \textit{(1) relationship between LPIPS and DSD}, and \textit{(2) relationship between LPIPS and SSIM}.
Next, we discuss the two relationships with examples.

\begin{figure}[t]
  \centering
  \includegraphics[width=\columnwidth]{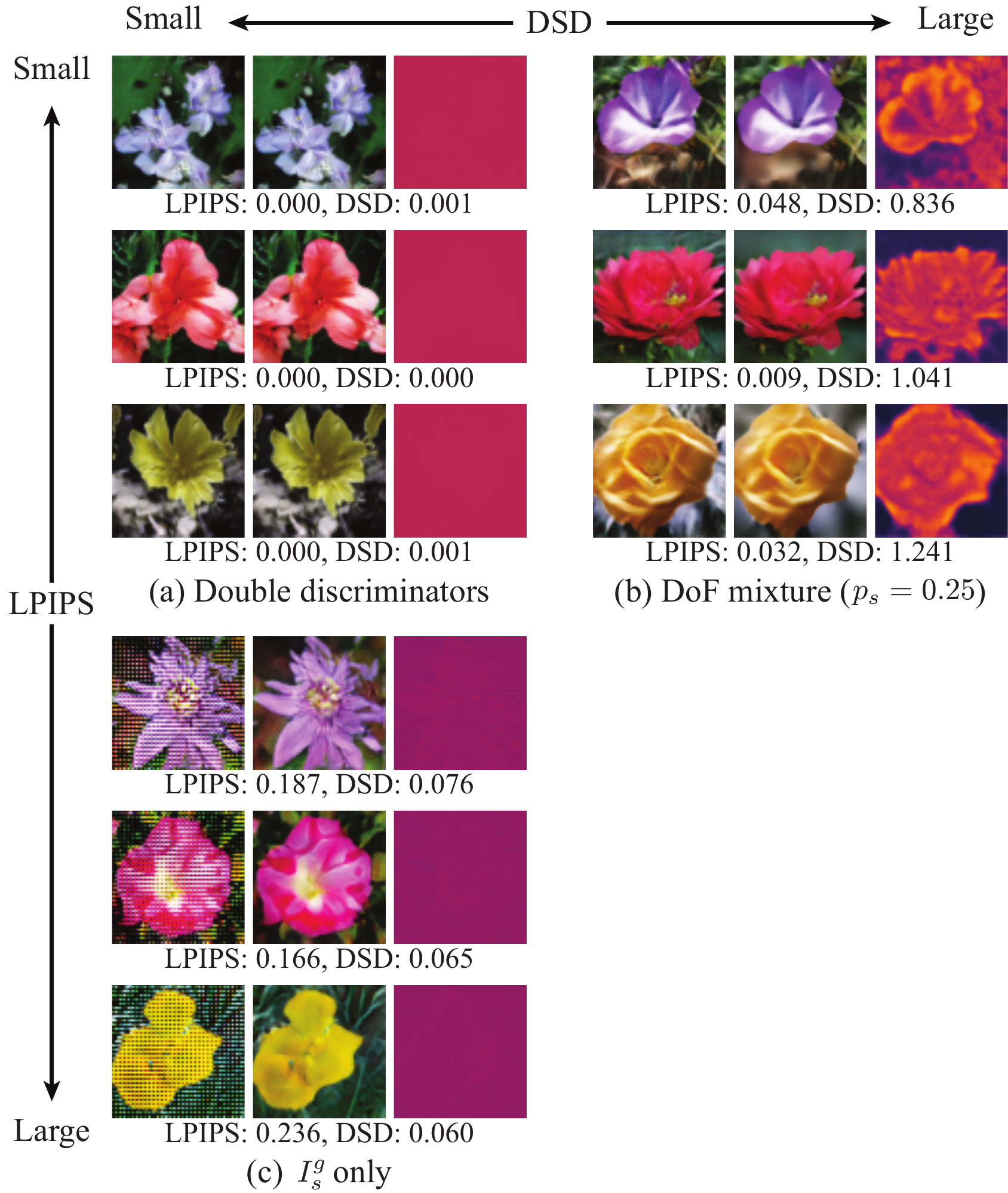}
  \caption{\textbf{Relationship between LPIPS and DSD.}
    In each block (a--c), the three images in the rows represent the generated deep DoF image (i.e., $I_d^g$), shallow DoF image (i.e., $I_s^g$), and depth (i.e., $D^g$), from left to right.
    The three images in the columns were generated from different noise $z$.
    For ease of viewing, the depth was normalized using the same values across all results.
    The models were trained using Oxford Flowers.}
  \label{fig:relationship_lpips_dsd}
  \vspace{-2mm}
\end{figure}

\smallskip\textit{(1) Relationship between LPIPS and DSD.}
Figure~\ref{fig:relationship_lpips_dsd} shows the relationship between LPIPS and DSD.
We selected three representative models based on the results in Table~\ref{tab:ablation_dof_mixture_learning_ex}(a):
\textit{(a) AR-GAN with double discriminators (No.~7)}, for which the LPIPS and DSD are small,
\textit{(b) AR-GAN with DoF mixture learning at $p_s = 0.25$ (No.~4)}, for which the LPIPS is small and the DSD is large, and
\textit{(c) AR-GAN with $I_s^g$ only (No.~1)}, for which the LPIPS is large and the DSD is small.

As shown in this figure, when the DSD is small (a, c), the model fails to learn the depth (i.e., the depth is almost constant).
By contrast, when the DSD is large (b), the model succeeds in learning the meaningful depth.
Furthermore, when the LPIPS is small (a, b), the content is preserved between the deep and shallow DoF images.
By contrast, when the LPIPS is large (c), the deep DoF image is corrupted.
As described, LPIPS and DSD reflect the success and failure of the learning of the depth and DoF effect reasonably well.
Therefore, we believe that they are reasonable metrics for this task.

\begin{figure}[t]
  \centering
  \includegraphics[width=\columnwidth]{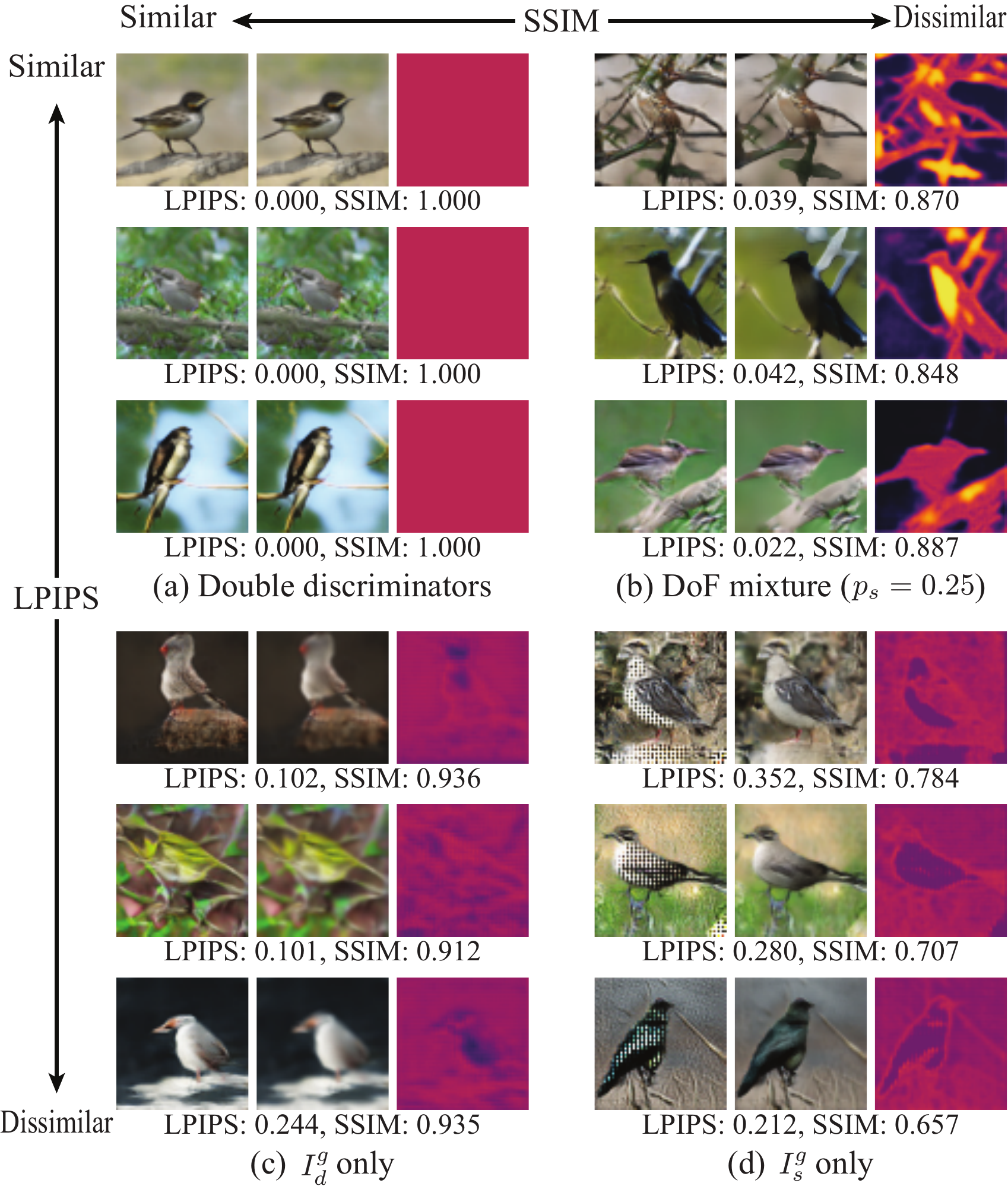}
  \caption{\textbf{Relationship between LPIPS and SSIM.}
    In each block (a--d), the three images in the rows represent the generated deep DoF image (i.e., $I_d^g$), shallow DoF image (i.e., $I_s^g$), and depth (i.e., $D^g$), from left to right.
    The three images in the columns were generated from different noise $z$.
    For ease of viewing, the depth was normalized using the same values across all results.
    The models were trained using CUB-200-2011.}
  \label{fig:relationship_lpips_ssim}
  \vspace{-2mm}
\end{figure}

\smallskip\textit{(2) Relationship between LPIPS and SSIM.}
Figure~\ref{fig:relationship_lpips_ssim} shows the relationship between LPIPS and SSIM.
We selected four representative models based on the results in Table~\ref{tab:ablation_dof_mixture_learning_ex}(b):
\textit{(a) AR-GAN with double discriminators (No.~7)}, in which deep and shallow DoF images are similar in terms of both LPIPS and SSIM,
\textit{(b) AR-GAN with DoF mixture learning at $p_s = 0.25$ (No.~4)}, in which deep and shallow DoF images are similar in terms of LPIPS but dissimilar in terms of SSIM,
\textit{(c) AR-GAN with $I_d^g$ only (No.~5)}, in which deep and shallow DoF images are similar in terms of SSIM but dissimilar in terms of LPIPS, and
\textit{(d) AR-GAN with $I_s^g$ only (No.~1)}, in which deep and shallow DoF images are dissimilar in terms of both LPIPS and SSIM.

As shown in this figure, LPIPS is sensitive to overall blur (c), in which both the main object and its surroundings are blurred.\footnote{This tendency has also been observed in the original study on LPIPS~\cite{RZhangCVPR2018}.}
However, it is insensitive to center focus blur (b), in which the main object in focus is not blurred but the surrounding area is blurred.
By contrast, SSIM is sensitive to the center focus blur (b) but insensitive to the overall blur (c).
This is because LPIPS measures the distance based on the classifier that is sensitive to the difference in the main target but insensitive to the differences in the irrelevant surroundings.
Our objective is to learn the natural DoF effect, in which the main target and the surroundings tend to be unchanged and changed, respectively, before and after defocusing.
To achieve this objective, we believe that LPIPS is more appropriate than SSIM as a metric that measures the unexpected dissimilarity caused by factors other than the DoF effect.

\begin{table*}
  \newcommand{\spm}[1]{{\tiny$\pm#1$}}
  \setlength{\tabcolsep}{3pt}
  \centering
  \scriptsize
  \begin{subtable}{\textwidth}
    \caption{Oxford Flowers}
    \vspace{-2mm}
    \begin{tabularx}{\textwidth}{cclCCCCCC}
    \toprule
    No. & \multicolumn{2}{c}{Sampling method} & KID$\times 10^3$$\downarrow$ & KID$_{I_s^g}$$\times 10^3$$\downarrow$ & KID$_{I_d^g}$$\times 10^3$$\downarrow$ & SIDE$\times 10^2$$\downarrow$ & LPIPS$\downarrow$ & DSD$\uparrow$
    \\ \midrule
    1 & $I_s^g$ only & $B(1, 1)$
    & 12.36 \spm{0.59}
    & 12.46 \spm{0.58}
    & 293.56 \spm{44.92}
    & 5.48 \spm{0.20}
    & 0.229 \spm{0.027}
    & 0.157 \spm{0.063}
    \\
    2 & Mixture & $B(1, 0.75)$
    & 10.97 \spm{0.26}
    & 13.74 \spm{0.47}
    & 26.50 \spm{1.59}
    & 4.81 \spm{0.06}
    & 0.023 \spm{0.001}
    & 0.657 \spm{0.006}
    \\
    3 & Mixture & $B(1, 0.5)$
    & 10.69 \spm{0.48}
    & 18.98 \spm{1.24}
    & 17.58 \spm{0.53}
    & 4.65 \spm{0.05}
    & 0.022 \spm{0.000}
    & 0.771 \spm{0.022}
    \\
    4 & Mixture & $B(1, 0.25)$
    & 11.23 \spm{0.36}
    & 30.61 \spm{2.99}
    & 12.91 \spm{0.38}
    & 4.46 \spm{0.03}
    & 0.028 \spm{0.001}
    & 1.007 \spm{0.025}
    \\
    5 & $I_d^g$ only & $B(1, 0)$
    & 11.58 \spm{0.37}
    & 36.63 \spm{5.11}
    & 11.51 \spm{0.41}
    & 4.56 \spm{0.20}
    & 0.113 \spm{0.013}
    & 0.446 \spm{0.065}
    \\ \midrule
    6 & Mixture & $U(0, 1)$
    & 10.67 \spm{0.16}
    & 21.14 \spm{2.49}
    & 29.72 \spm{0.80}
    & 4.31 \spm{0.11}
    & 0.040 \spm{0.002}
    & 0.857 \spm{0.118}
    \\ \bottomrule    
  \end{tabularx}
\end{subtable}
\begin{subtable}{\textwidth}
  \vspace{1mm}
  \caption{CUB-200-2011}
  \vspace{-2mm}
  \begin{tabularx}{\textwidth}{cclCCCCCC}
    \toprule
    No. & \multicolumn{2}{c}{Sampling method} & KID$\times 10^3$$\downarrow$ & KID$_{I_s^g}$$\times 10^3$$\downarrow$ & KID$_{I_d^g}$$\times 10^3$$\downarrow$ & SIDE$\times 10^2$$\downarrow$ & LPIPS$\downarrow$ & DSD$\uparrow$
    \\ \midrule
    1 & $I_s^g$ only & $B(1, 1)$
    & 13.62 \spm{0.53}
    & 13.71 \spm{0.58}
    & 72.63 \spm{25.71}
    & 4.63 \spm{0.50}
    & 0.125 \spm{0.037}
    & 0.354 \spm{0.021}
    \\
    2 & Mixture & $B(1, 0.75)$
    & 12.68 \spm{0.61}
    & 12.30 \spm{0.45}
    & 22.64 \spm{0.37}
    & 3.75 \spm{0.08}
    & 0.037 \spm{0.003}
    & 0.748 \spm{0.072}
    \\
    3 & Mixture & $B(1, 0.5)$
    & 13.14 \spm{0.03}
    & 13.20 \spm{0.43}
    & 18.15 \spm{0.07}
    & 3.55 \spm{0.02}
    & 0.043 \spm{0.003}
    & 0.959 \spm{0.075}
    \\
    4 & Mixture & $B(1, 0.25)$
    & 14.30 \spm{0.56}
    & 18.28 \spm{0.73}
    & 16.19 \spm{0.63}
    & 3.58 \spm{0.04}
    & 0.059 \spm{0.002}
    & 1.175 \spm{0.017}
    \\
    5 & $I_d^g$ only & $B(1, 0)$
    & 14.58 \spm{0.56}
    & 28.63 \spm{2.14}
    & 14.85 \spm{0.72}
    & 5.94 \spm{0.70}
    & 0.115 \spm{0.019}
    & 0.193 \spm{0.012}
    \\ \midrule
    6 & Mixture & $U(0, 1)$
    & 12.43 \spm{0.92}
    & 13.38 \spm{0.70}
    & 25.60 \spm{1.89}
    & 3.62 \spm{0.04}
    & 0.072 \spm{0.006}
    & 1.142 \spm{0.038}
    \\ \bottomrule
  \end{tabularx}
\end{subtable}
\begin{subtable}{\textwidth}
  \vspace{1mm}
  \caption{FFHQ}
  \vspace{-2mm}
\begin{tabularx}{\textwidth}{cclCCCCCC}
    \toprule
    No. & \multicolumn{2}{c}{Sampling method} & KID$\times 10^3$$\downarrow$ & KID$_{I_s^g}$$\times 10^3$$\downarrow$ & KID$_{I_d^g}$$\times 10^3$$\downarrow$ & SIDE$\times 10^2$$\downarrow$ & LPIPS$\downarrow$ & DSD$\uparrow$
    \\ \midrule
    1 & $I_s^g$ only & $B(1, 1)$
    & 5.75 \spm{0.44}
    & 5.82 \spm{0.48}
    & 74.36 \spm{7.41}
    & 6.00 \spm{0.35}
    & 0.097 \spm{0.011}
    & 0.296 \spm{0.018}
    \\
    2 & Mixture & $B(1, 0.75)$
    & 5.67 \spm{0.23}
    & 6.05 \spm{0.14}
    & 14.50 \spm{0.65}
    & 4.38 \spm{0.10}
    & 0.009 \spm{0.001}
    & 0.757 \spm{0.177}
    \\
    3 & Mixture & $B(1, 0.5)$
    & 5.75 \spm{0.19}
    & 7.53 \spm{0.43}
    & 9.34 \spm{0.22}
    & 4.21 \spm{0.15}
    & 0.009 \spm{0.001}
    & 0.769 \spm{0.119}
    \\
    4 & Mixture & $B(1, 0.25)$
    & 6.17 \spm{0.08}
    & 11.55 \spm{0.70}
    & 7.53 \spm{0.41}
    & 4.68 \spm{0.33}
    & 0.010 \spm{0.001}
    & 0.583 \spm{0.071}
    \\
    5 & $I_d^g$ only & $B(1, 0)$
    & 6.85 \spm{0.13}
    & 10.16 \spm{0.29}
    & 6.94 \spm{0.10}
    & 4.77 \spm{0.13}
    & 0.028 \spm{0.006}
    & 0.202 \spm{0.003}
    \\ \midrule
    6 & Mixture & $U(0, 1)$
    & 5.86 \spm{0.06}
    & 7.30 \spm{0.02}
    & 17.86 \spm{0.25}
    & 4.49 \spm{0.09}
    & 0.020 \spm{0.001}
    & 1.130 \spm{0.085}
    \\ \bottomrule
  \end{tabularx}
\end{subtable}
\vspace{0mm}
  \caption{\textbf{Comparison of KID$\times 10^3$$\downarrow$, KID$_{I_s^g}$$\times 10^3$$\downarrow$, KID$_{I_d^g}$$\times 10^3$$\downarrow$, SIDE$\times 10^2$$\downarrow$, LPIPS$\downarrow$, and DSD$\uparrow$ among AR-GANs with different sampling methods.}}
  \label{tab:analysis_sampling}
  \vspace{-2mm}
\end{table*}

\subsection{Further analysis of DoF mixture learning}
\label{subsec:further_analysis_dof_mixture_learning}

\subsubsection{Effect of sampling methods}
\label{subsubsec:effect_of_sampling_methods}

As described in Section~\ref{subsec:dof_mixture_learning}, we sampled $s$ (Equation~\ref{eqn:ar-gan}) from a binomial distribution, which is $p(s) = B(1, p_s)$.
As demonstrated in the experiments in Section~\ref{sec:experiments}, this setting works reasonably well.
However, as a further analysis, in this subsection, we evaluate the performance when $s$ is sampled from a different distribution, particularly a uniform distribution, which is $p(s) = U(0, 1)$.
  
\smallskip\noindent\textbf{Results.}
Table~\ref{tab:analysis_sampling} summarizes the comparison of KID, KID$_{I_s^g}$, KID$_{I_d^g}$, SIDE, LPIPS, and DSD among AR-GANs with different sampling methods.
When a binomial distribution is used (i.e., $s \in \{ 0, 1\}$), the relationship between the deep and shallow DoF images is discretely represented.
By contrast, when a uniform distribution is used (i.e., $s \in [0, 1]$), the relationship is continuously represented.
In the latter case, deep and shallow DoF images are represented continuously, and their total distribution is encouraged to represent the overall real distribution.
Therefore, $I_d^g$ ($s = 0$) represents the deepest DoF image, and $I_s^g$ ($s = 1$) represents the most shallow DoF image among the various DoF images.
Consequently, the difference between $I_d^g$ and $I_s^g$ becomes larger compared with the use of a binomial distribution.
Therefore, in Table~\ref{tab:analysis_sampling}, the LPIPS for $U(0, 1)$ (No.~6) is larger than those for $B(1, p_s)$ at $p_s \in \{ 0.25, 0.5, 0.75\}$ (Nos.~2--4) and the DSD for $U(0, 1)$ (No.~6) is comparable with those for $B(1, p_s)$, which achieved the largest DSD among Nos.~2--4.
Because of this characteristic difference, there are advantages and disadvantages for both a uniform distribution and a binomial distribution.
The choice of sampling methods is therefore task-specific.

\begin{table*}
  \newcommand{\spm}[1]{{\tiny$\pm#1$}}
  \setlength{\tabcolsep}{3pt}
  \centering
  \scriptsize
  \begin{subtable}{\textwidth}
    \caption{Oxford Flowers}
    \vspace{-2mm}
    \begin{tabularx}{\textwidth}{ccCCCCCC}
    \toprule
    No. & Distance metric & KID$\times 10^3$$\downarrow$ & KID$_{I_s^g}$$\times 10^3$$\downarrow$ & KID$_{I_d^g}$$\times 10^3$$\downarrow$ & SIDE$\times 10^2$$\downarrow$ & LPIPS$\downarrow$ & DSD$\uparrow$
    \\ \midrule
    1 & L1
    & 11.66 \spm{0.72}
    & 11.81 \spm{0.74}
    & 47.59 \spm{2.54}
    & 5.26 \spm{0.65}
    & 0.033 \spm{0.001}
    & 0.387 \spm{0.116}
    \\
    2 & Perceptual loss
    & 11.18 \spm{0.52}
    & 11.33 \spm{0.58}
    & 40.35 \spm{0.56}
    & 4.70 \spm{0.06}
    & 0.028 \spm{0.001}
    & 0.589 \spm{0.041}
    \\ \bottomrule    
  \end{tabularx}
\end{subtable}
\begin{subtable}{\textwidth}
  \vspace{1mm}
  \caption{CUB-200-2011}
  \vspace{-2mm}
  \begin{tabularx}{\textwidth}{ccCCCCCC}
    \toprule
    No. & Distance metrics & KID$\times 10^3$$\downarrow$ & KID$_{I_s^g}$$\times 10^3$$\downarrow$ & KID$_{I_d^g}$$\times 10^3$$\downarrow$ & SIDE$\times 10^2$$\downarrow$ & LPIPS$\downarrow$ & DSD$\uparrow$
    \\ \midrule
    1 & L1
    & 12.54 \spm{0.32}
    & 12.59 \spm{0.45}
    & 33.70 \spm{0.84}
    & 5.75 \spm{1.26}
    & 0.042 \spm{0.001}
    & 0.725 \spm{0.195}
    \\
    2 & Perceptual loss
    & 12.78 \spm{0.21}
    & 12.82 \spm{0.23}
    & 36.16 \spm{2.34}
    & 3.75 \spm{0.07}
    & 0.054 \spm{0.003}
    & 0.754 \spm{0.053}
    \\ \bottomrule
  \end{tabularx}
\end{subtable}
\begin{subtable}{\textwidth}
  \vspace{1mm}
  \caption{FFHQ}
  \vspace{-2mm}
\begin{tabularx}{\textwidth}{ccCCCCCC}
    \toprule
    No. & Distance metric & KID$\times 10^3$$\downarrow$ & KID$_{I_s^g}$$\times 10^3$$\downarrow$ & KID$_{I_d^g}$$\times 10^3$$\downarrow$ & SIDE$\times 10^2$$\downarrow$ & LPIPS$\downarrow$ & DSD$\uparrow$
    \\ \midrule
    1 & L1
    & 5.82 \spm{0.21}
    & 5.83 \spm{0.23}
    & 24.89 \spm{2.70}
    & 4.82 \spm{0.09}
    & 0.015 \spm{0.004}
    & 0.466 \spm{0.045}
    \\
    2 & Perceptual loss
    & 5.64 \spm{0.40}
    & 5.66 \spm{0.42}
    & 39.37 \spm{1.46}
    & 5.31 \spm{0.07}
    & 0.024 \spm{0.002}
    & 0.439 \spm{0.010}
    \\ \bottomrule
  \end{tabularx}
\end{subtable}
\vspace{0mm}
\caption{\textbf{Comparison of KID$\times 10^3$$\downarrow$, KID$_{I_s^g}$$\times 10^3$$\downarrow$, KID$_{I_d^g}$$\times 10^3$$\downarrow$, SIDE$\times 10^2$$\downarrow$, LPIPS$\downarrow$, and DSD$\uparrow$ between AR-GANs using different distance metrics.}
  Refer to Table~\ref{tab:ablation_dof_mixture_learning_ex} for a comparison with the results of other learning methods.}
  \label{tab:analysis_distance_metrics}
  \vspace{-2mm}
\end{table*}

\begin{table*}
  \newcommand{\spm}[1]{{\tiny$\pm#1$}}
  \setlength{\tabcolsep}{2pt}
  \centering
  \scriptsize
  \begin{subtable}{\textwidth}
    \caption{Oxford Flowers}
    \vspace{-2mm}
    \begin{tabularx}{\textwidth}{cclCCCCCC}
      \toprule
      No. & \multicolumn{2}{c}{Usage of prior} & KID$\times 10^3$$\downarrow$ & KID$_{I_s^g}$$\times 10^3$$\downarrow$ & KID$_{I_d^g}$$\times 10^3$$\downarrow$ & SIDE$\times 10^2$$\downarrow$ & LPIPS$\downarrow$ & DSD$\uparrow$
      \\ \midrule
      1 & W/ $D_p$ & at beginning
      & 11.23 \spm{0.36}
      & 30.61 \spm{2.99}
      & 12.91 \spm{0.38}
      & 4.46 \spm{0.03} 
      & 0.028 \spm{0.001}
      & 1.007 \spm{0.025}
      \\
      2 & W/ $D_p$ & until end
      & 10.98 \spm{0.27}
      & 29.33 \spm{0.52}
      & 12.70 \spm{0.50}
      & 4.54 \spm{0.01}
      & 0.019 \spm{0.001}
      & 0.259 \spm{0.001}
      \\
      3 & W/o $D_p$ &
      & 10.69 \spm{0.24}
      & 27.79 \spm{2.50}
      & 12.60 \spm{0.23}
      & 6.78 \spm{1.58} 
      & 0.026 \spm{0.002}
      & 0.915 \spm{0.137}
      \\ \bottomrule
    \end{tabularx}
  \end{subtable}
  \begin{subtable}{\textwidth}
    \vspace{1mm}
    \caption{CUB-200-2011}
    \vspace{-2mm}
    \begin{tabularx}{\textwidth}{cclCCCCCC}
      \toprule
      No. & \multicolumn{2}{c}{Usage of prior} & KID$\times 10^3$$\downarrow$ & KID$_{I_s^g}$$\times 10^3$$\downarrow$ & KID$_{I_d^g}$$\times 10^3$$\downarrow$ & SIDE$\times 10^2$$\downarrow$ & LPIPS$\downarrow$ & DSD$\uparrow$
      \\ \midrule
      1 & W/ $D_p$ & at beginning
      & 14.30 \spm{0.56}
      & 18.28 \spm{0.73}
      & 16.19 \spm{0.63}
      & 3.58 \spm{0.04}
      & 0.059 \spm{0.002}
      & 1.175 \spm{0.017}
      \\
      2 & W/ $D_p$ & until end
      & 13.96 \spm{0.76}
      & 16.71 \spm{1.06}
      & 15.11 \spm{0.95}
      & 5.29 \spm{0.02}
      & 0.034 \spm{0.000}
      & 0.250 \spm{0.001}
      \\
      3 & W/o $D_p$ &
      & 13.96 \spm{0.63}
      & 17.40 \spm{1.01}
      & 16.01 \spm{0.66}
      & 4.86 \spm{1.84}
      & 0.062 \spm{0.004}
      & 1.183 \spm{0.059}
      \\ \bottomrule
    \end{tabularx}
  \end{subtable}
  \begin{subtable}{\textwidth}
    \vspace{1mm}
    \caption{FFHQ}
    \vspace{-2mm}
    \begin{tabularx}{\textwidth}{cclCCCCCC}
      \toprule
      No. & \multicolumn{2}{c}{Usage of prior} & KID$\times 10^3$$\downarrow$ & KID$_{I_s^g}$$\times 10^3$$\downarrow$ & KID$_{I_d^g}$$\times 10^3$$\downarrow$ & SIDE$\times 10^2$$\downarrow$ & LPIPS$\downarrow$ & DSD$\uparrow$
      \\ \midrule
      1 & W/ $D_p$ & at beginning
      & 5.75 \spm{0.19}
      & 7.53 \spm{0.43}
      & 9.34 \spm{0.22}
      & 4.21 \spm{0.15}
      & 0.009 \spm{0.001}
      & 0.769 \spm{0.119}
      \\
      2 & W/ $D_p$ & until end
      & 5.67 \spm{0.17}
      & 7.98 \spm{0.36}
      & 9.06 \spm{0.22}
      & 4.31 \spm{0.04}
      & 0.008 \spm{0.001}
      & 0.271 \spm{0.004}
      \\
      3 & W/o $D_p$ &
      & 5.72 \spm{0.10}
      & 7.78 \spm{0.51}
      & 9.88 \spm{0.32}
      & 6.70 \spm{1.88}
      & 0.009 \spm{0.001}
      & 0.851 \spm{0.057}
      \\ \bottomrule
    \end{tabularx}
  \end{subtable}
  \vspace{0mm}
  \caption{\textbf{Comparison of KID$\times 10^3$$\downarrow$, KID$_{I_s^g}$$\times 10^3$$\downarrow$, KID$_{I_d^g}$$\times 10^3$$\downarrow$, SIDE$\times 10^2$$\downarrow$, LPIPS$\downarrow$, and DSD$\uparrow$ among AR-GANs with and without center focus prior.}
    The duration of the center focus prior usage changes between W/ $D_p$ at the beginning (No.~1) and W/ $D_p$ until the end (No.~2).
    This is an extended version of Table~\ref{tab:ablation_center_focus_prior}.}
  \label{tab:ablation_center_focus_prior_ex}
  \vspace{-2mm}
\end{table*}

\begin{figure*}[t]
  \centering
  \includegraphics[width=0.75\textwidth]{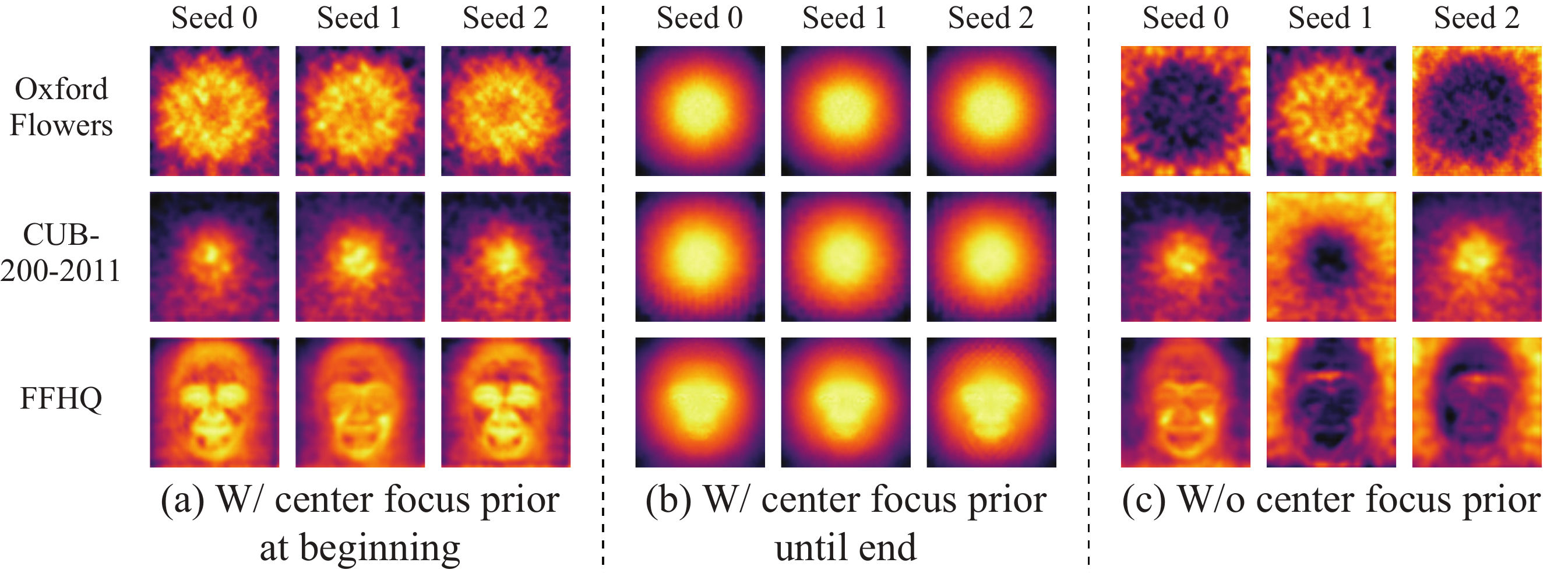}
  \caption{\textbf{Comparison of AD with and without $D_p$.}
    This is an extended version of Figure~\ref{fig:comparison_ad}.
    Light color indicates the foreground.}
  \label{fig:comparison_ad_ex}
  \vspace{-2mm}
\end{figure*}

\subsubsection{Comparison with another alternative learning}
\label{subsubsec:comparison_with_another_alternative_learning}

In Section~\ref{subsubsec:ablation_study_dof_mixture_learning}, we tested \textit{L1} and \textit{double discriminators} as possible alternatives to DoF mixture learning.
As a further alternative, we test \textit{perceptual loss}~\cite{JJohnsonECCV2016,ADosovitskiyNIPS2016}, which is a variant of explicit distance metrics and measures the distance between two images in the CNN feature space.
Recent studies~\cite{ADosovitskiyNIPS2016,CLedigCVPR2017} have shown that perceptual loss is more useful than pixel-level metrics (e.g., L1 and L2) when aiming to improve visual fidelity.

\smallskip\noindent\textbf{Results.}
Table~\ref{tab:analysis_distance_metrics} summarizes the comparison of KID, KID$_{I_s^g}$, KID$_{I_d^g}$, SIDE, LPIPS, and DSD between AR-GANs using different distance metrics.
We found that even when using perceptual loss (No.~2), explicit distance metric-based methods (Nos.~1 and 2) find it difficult to improve both the LPIPS and DSD simultaneously.
Furthermore, they also find it difficult to improve KID$_{I_d^g}$.
It is noteworthy that DoF mixture learning can address these difficulties, as listed in Table~\ref{tab:ablation_dof_mixture_learning_ex} and discussed in Appendix~\ref{subsubsec:evaluation_on_additional_metrics}.
These results indicate that the aforementioned difficulties are common in explicit distance metric-based methods, including L1 and perceptual loss.

\subsection{Further analysis of center focus prior}
\label{subsec:further_analysis_center_focus_prior}

\subsubsection{Effect of duration of using center focus prior}
\label{subsubsec:effect_of_duration_of_using_center_focus_prior}

In Section~\ref{subsubsec:ablation_study_center_focus_prior}, we used the center focus prior only at the beginning of training to mitigate the negative effect caused by the gap between the real depth and the predefined center focus prior.
To validate this strategy, we evaluated the performance of the model that used the center focus prior until the end of training.

\smallskip\noindent\textbf{Results.}
Table~\ref{tab:ablation_center_focus_prior_ex} summarizes the comparison of KID, KID$_{I_s^g}$, KID$_{I_d^g}$, SIDE, LPIPS, and DSD among AR-GANs with and without center focus prior.
Specifically, we changed the duration of the center focus prior usage between \textit{w/ $D_p$ at beginning} (No.~1) and \textit{w/ $D_p$ until end} (No.~2).
As discussed in Section~\ref{subsubsec:ablation_study_center_focus_prior}, when $D_p$ is used only at the beginning of the training (No.~1), the score differences between the models with and without $D_p$ (Nos.~1 and 3) are relatively small.
However, when $D_p$ is used until the end of the training (No.~2), the DSD is significantly smaller than the results without $D_p$ (No.~3).
These results indicate that when $D_p$ is used until the end of the training, the depth learning is disturbed, but when its usage is stopped in the early phase of training, learning is not prevented.

As further evidence, we show a comparison of the average depth (AD) in Figure~\ref{fig:comparison_ad_ex}.
As shown here, when $D_p$ is used only at the beginning of the training (a), AD is deformed depending on the dataset.
However, when $D_p$ is used until the end of the training (b), AD is almost the same as the center focus prior shown in Figure~\ref{fig:center_focus_prior}(b).
These results also support the aforementioned statement that when $D_p$ is used until the end of the training, the depth learning is disturbed.

\clearpage

\section{Extended results}
\label{sec:extended_results}

\begin{figure*}[t]
  \centering
  \includegraphics[width=\textwidth]{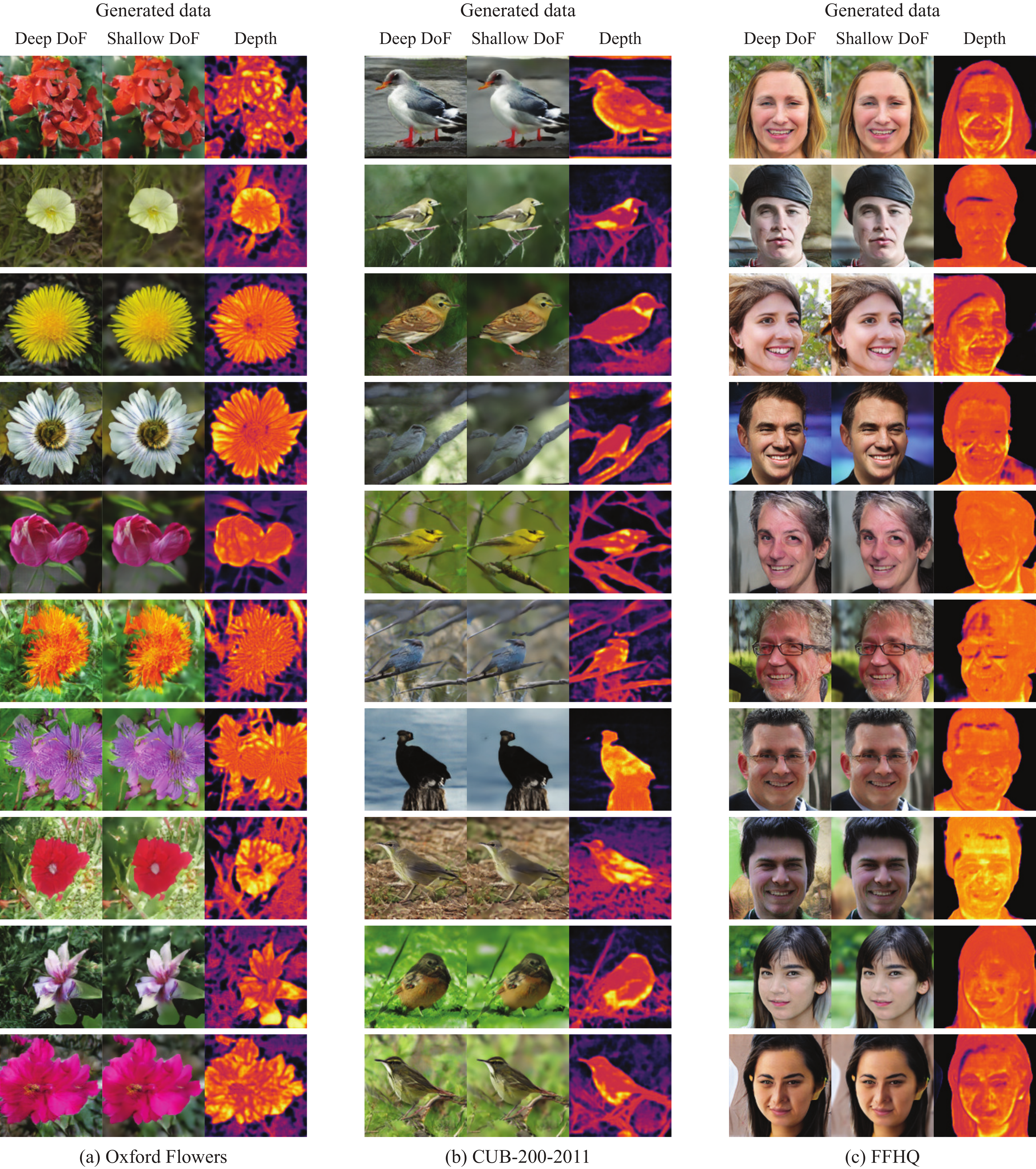}
  \caption{\textbf{Unsupervised learning of depth and DoF effect from unlabeled natural images.}
    This is an extended version of Figure~\ref{fig:concept}.
    Once trained, our model can synthesize tuples of deep and shallow DoF images and depths from random noise.
    Here, we show the results generated using AR-GANs trained with $128 \times 128$ images.
    The three images in each image block represent the generated deep DoF image (i.e., $I_d^g$), shallow DoF image (i.e., $I_s^g$), and depth (i.e., $D^g$), from left to right.
    The images in the columns are generated from different noise (i.e., $z$).}
  \label{fig:concept_ex}
\end{figure*}

\begin{figure*}[t]
  \centering
  \includegraphics[width=\textwidth]{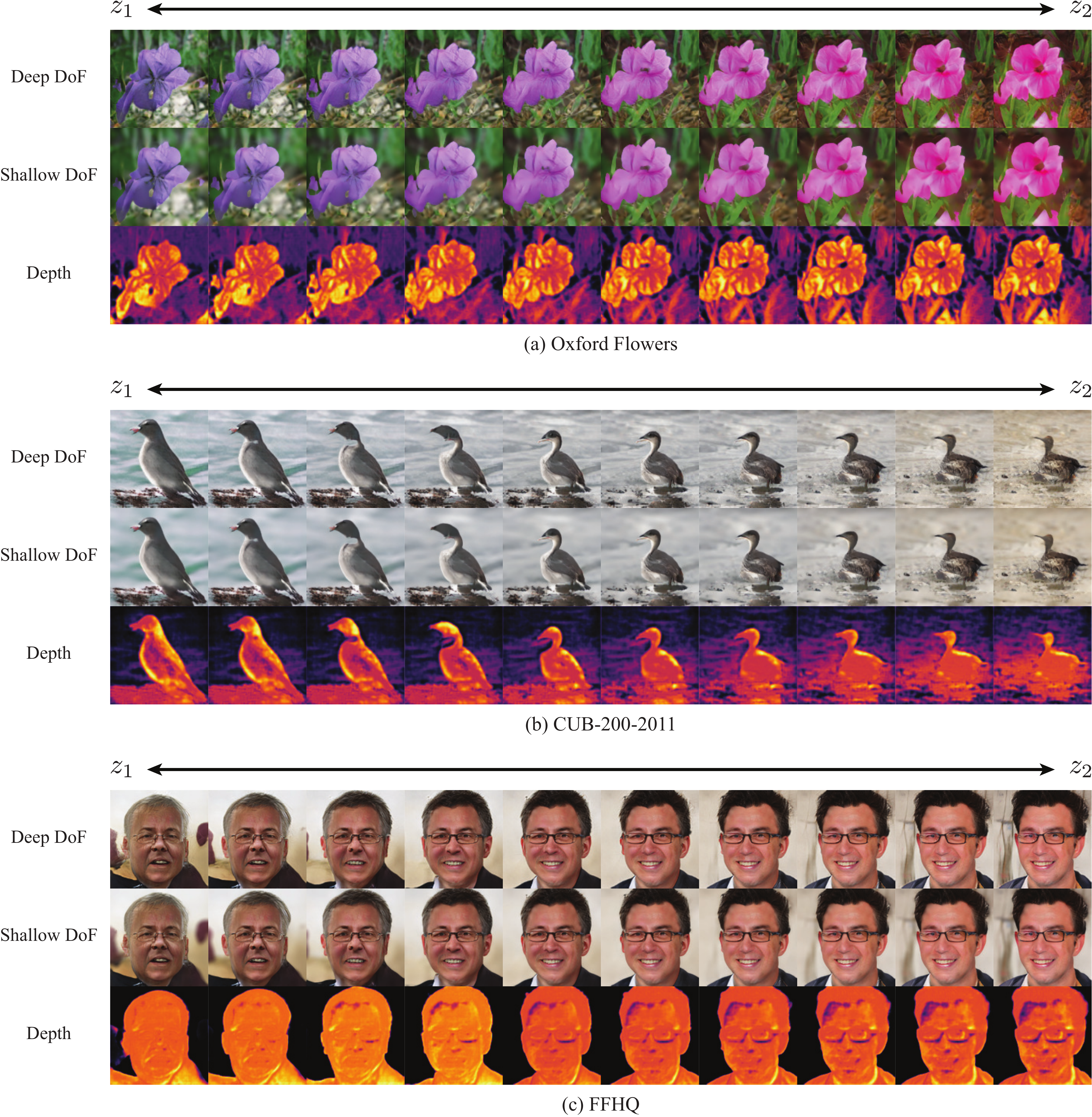}
  \caption{\textbf{Linear interpolation in the latent space of the AR-GAN generator.}
    This is an extended version of Figure~\ref{fig:concept}.
    In the horizontal direction, we linearly interpolated the noise $z \in [ z_1, z_2 ]$ in the latent space of the AR-GAN generator.
    The three images in the columns represent the deep DoF image (i.e., $I_d^g$), shallow DoF image (i.e., $I_s^g$), and depth (i.e., $D^g$) that were generated from the corresponding noise $z$, from top to bottom.
    As shown here, AR-GANs can manipulate the contents of tuples of $(I_s^g, I_d^g, D^g)$ continuously according to $z$.}
  \label{fig:interpolation}
\end{figure*}

\begin{figure*}[t]
  \centering
  \includegraphics[height=0.9\textheight]{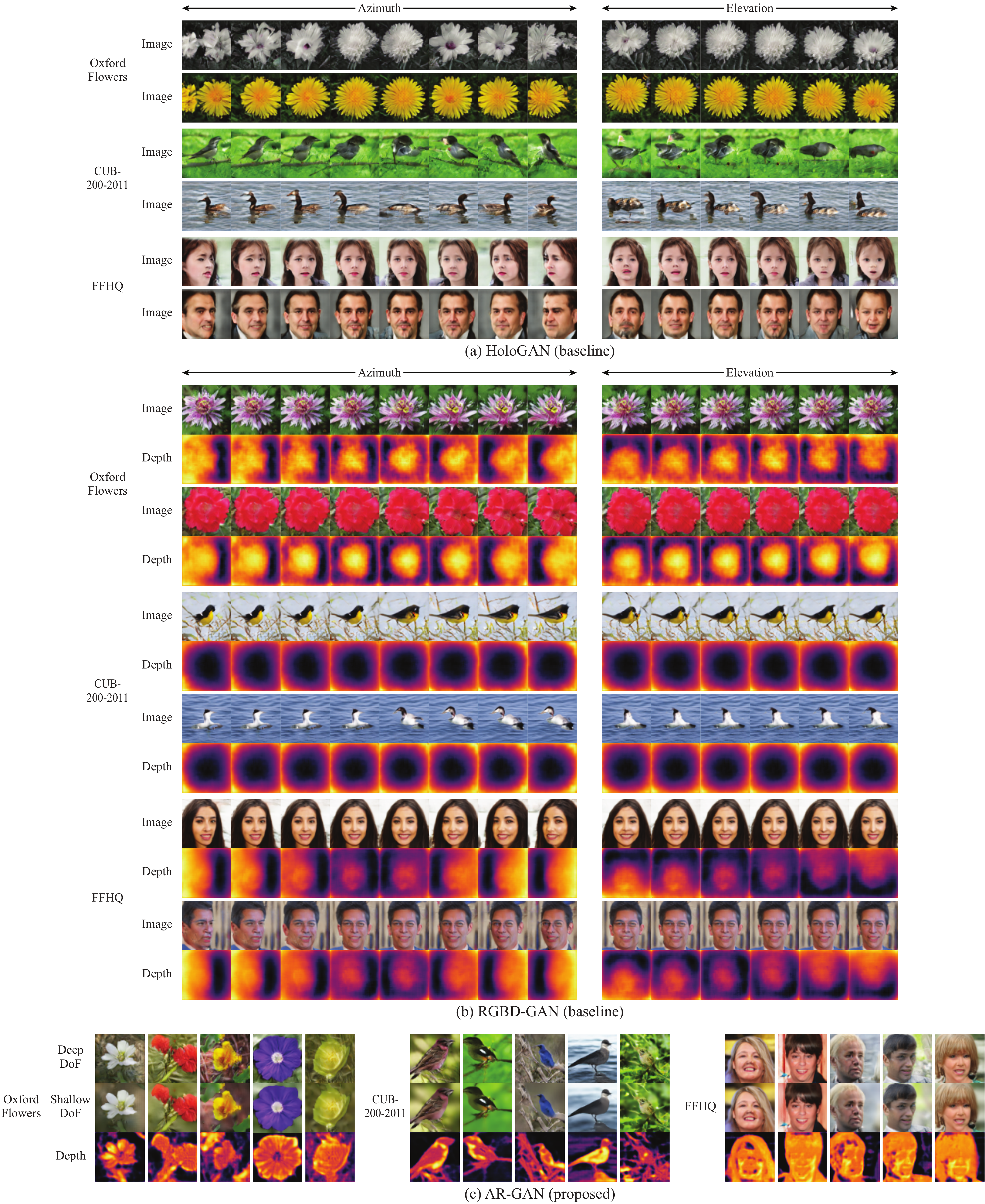}
  \caption{\textbf{Qualitative comparison among HoloGAN, RGBD-GAN, and AR-GAN.}
    This is an extended version of Figure~\ref{fig:comparison_gans}.
    HoloGAN generates images only, RGBD-GAN generates image and depth pairs, and AR-GAN generates tuples of deep and shallow DoF images and depths.
    The respective sets are shown in the same column in each image block.
    When viewpoint distributions are biased (i.e., Oxford Flowers and CUB-200-2011), HoloGAN and RGBD-GAN have difficulty in learning 3D representations (e.g., in (a), the bird direction does not change smoothly in HoloGAN on CUB-200-2011), whereas AR-GAN succeeds in learning the depth and DoF effect in all datasets.}
  \label{fig:comparison_gans_ex}
\end{figure*}

\begin{figure*}[t]
  \centering
  \includegraphics[height=0.8\textheight]{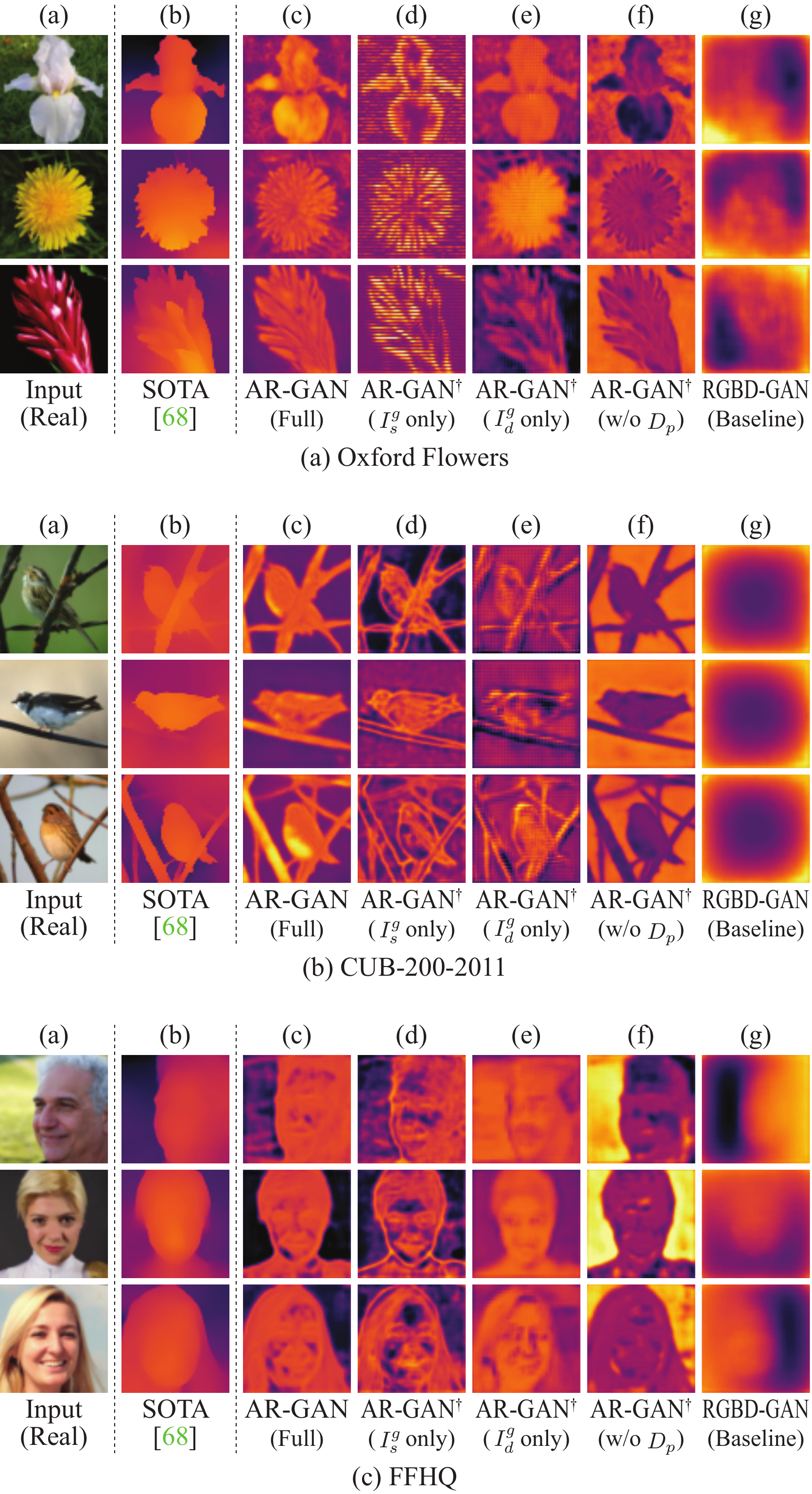}
  \caption{\textbf{Examples of predicted depths.}
    This is an extended version of Figure~\ref{fig:depth_prediction}.
    (b) Results obtained using a state-of-the-art (SOTA) monocular depth estimator~\cite{KXianCVPR2020}, which was trained using stereo pairs in an external dataset.
    (c--g) Results obtained in a fully unsupervised setting.
    Particularly in (c), AR-GAN with DoF mixture learning and center focus prior was used.
    In (d--f), $^{\dag}$ indicates the ablated model, and either DoF mixture learning or the center focus prior was ablated.
    In (g), RGBD-GAN was used.}
  \label{fig:depth_prediction_ex}
\end{figure*}

\begin{figure*}[t]
  \centering
  \includegraphics[width=\textwidth]{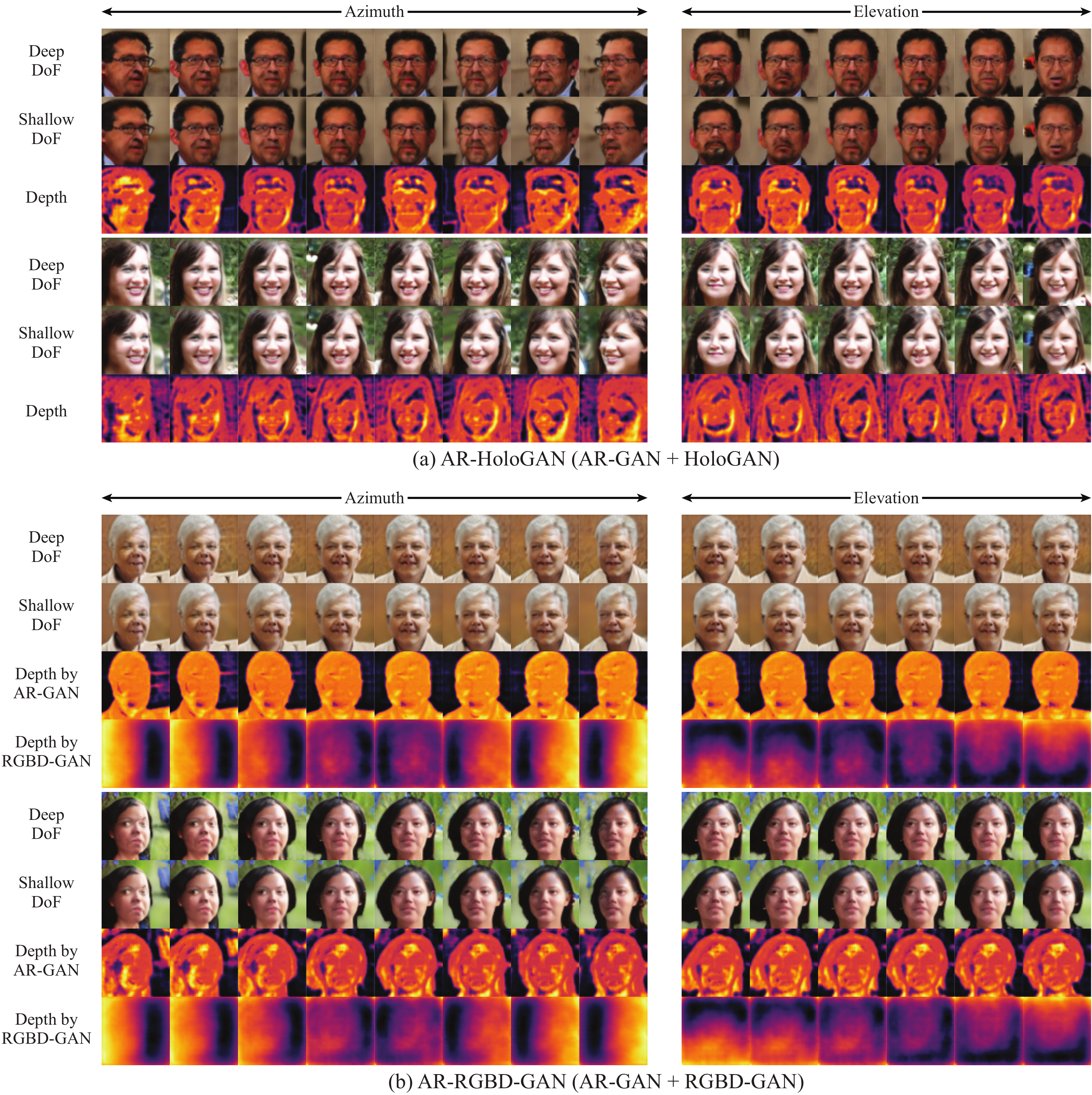}
  \caption{\textbf{Examples of data generated using AR-HoloGAN and AR-RGBD-GAN.}
    This is an extended version of Figure~\ref{fig:combination}.
    The viewpoint change in the horizontal direction is obtained by the HoloGAN/RGBD-GAN function, whereas the DoF change and depth in the vertical direction (in the top three rows in each image block) are obtained by the AR-GAN function.
    With regard to AR-RGBD-GAN, we also visualized the depth predicted by the RGBD-GAN function in the bottom row.
    Two types of depths (i.e., the depth predicted by AR-GAN and the depth predicted by RGBD-GAN) do not completely match because they are calculated based on different principles.
    The depth predicted by AR-GAN is based on \textit{focus cues}, whereas the depth predicted by RGBD-GAN is based on the \textit{consistency between viewpoints}.}
  \label{fig:combination_ex}
\end{figure*}

\begin{figure*}[t]
  \centering
  \includegraphics[height=0.85\textheight]{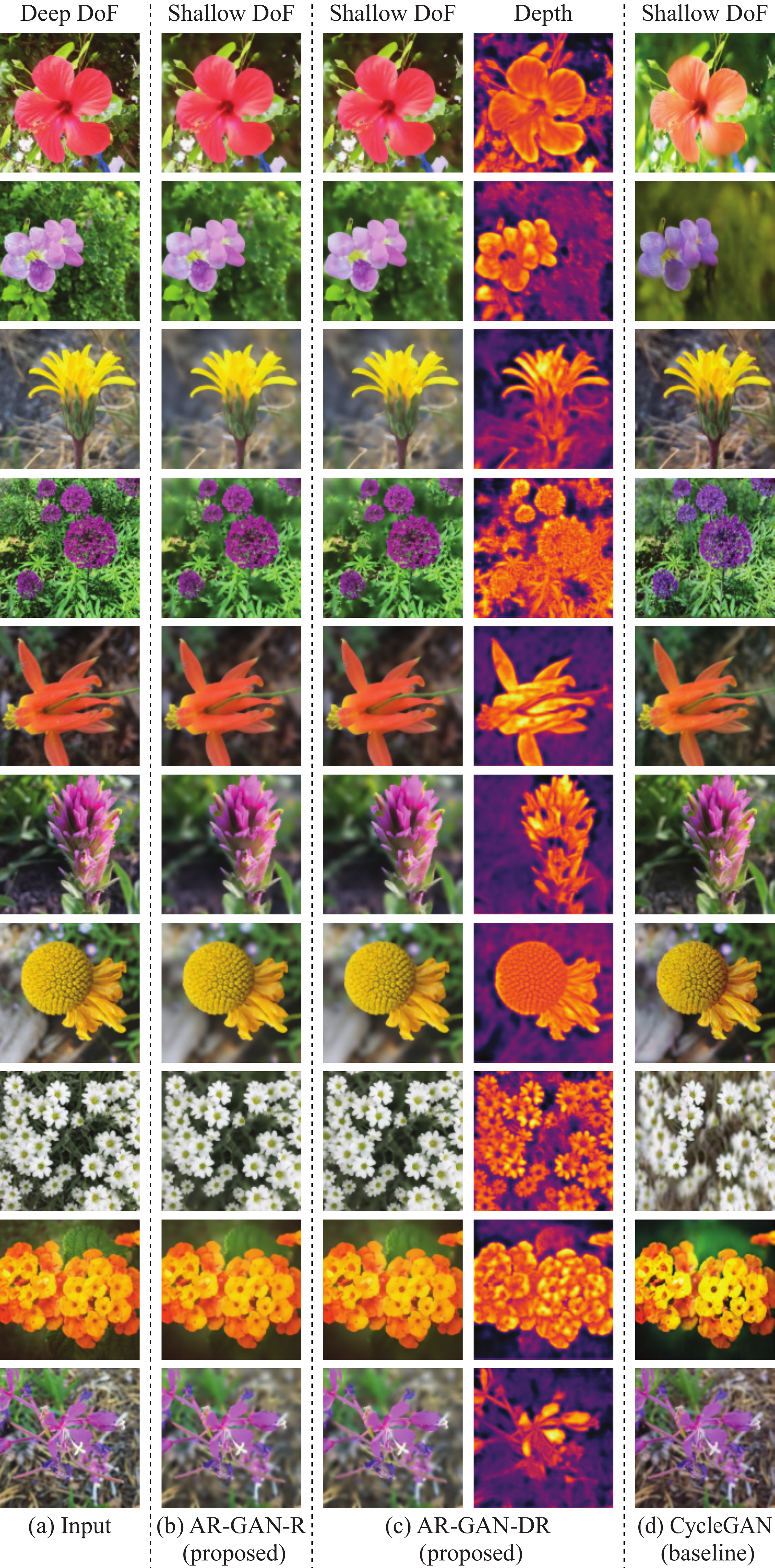}
  \caption{\textbf{Examples of shallow DoF rendering.}
    This is an extended version of Figure~\ref{fig:shallow_dof_rendering}.
    CycleGAN often yields unnecessary changes (e.g., color change in the first row or overblurring of the main object (i.e., the flower) in the eighth row), whereas AR-GAN-R and AR-GAN-DR do not.
    The generated shallow DoF images were almost the same for AR-GAN-R and AR-GAN-DR.
    The main difference between them is that AR-GAN-DR can estimate the depth simultaneously, whereas AR-GAN-R cannot.}
  \label{fig:shallow_dof_rendering_ex}
  \vspace{-3mm}
\end{figure*}

In this appendix, we provide extended versions of the figures presented in the main text.
Figure titles and the relationships with the figures presented in the main text are as follows:
\begin{itemize}
\item Figure~\ref{fig:concept_ex}:
  Unsupervised learning of depth and DoF effect from unlabeled natural images.
  This is an extended version of Figure~\ref{fig:concept}.
\item Figure~\ref{fig:interpolation}:
  Linear interpolation in the latent space of the AR-GAN generator.
  This is an extended version of Figure~\ref{fig:concept}.
\item Figure~\ref{fig:comparison_gans_ex}:
  Qualitative comparison among HoloGAN, RGBD-GAN, and AR-GAN.
  This is an extended version of Figure~\ref{fig:comparison_gans}.
\item Figure~\ref{fig:depth_prediction_ex}:
  Examples of predicted depths.
  This is an extended version of Figure~\ref{fig:depth_prediction}.
\item Figure~\ref{fig:combination_ex}:
  Examples of data generated using AR-HoloGAN and AR-RGBD-GAN.
  This is an extended version of Figure~\ref{fig:combination}.
\item Figure~\ref{fig:shallow_dof_rendering_ex}:
  Examples of shallow DoF rendering.
  This is an extended version of Figure~\ref{fig:shallow_dof_rendering}.
\end{itemize}

\clearpage
\section{Implementation details}
\label{sec:implementation_details}

In this appendix, we provide implementation details for the following items:
\begin{itemize}
\item Appendix~\ref{subsec:details_image_generation}:
  Implementation details of $64 \times 64$ image generation, which were used in the experiments in Sections~\ref{subsec:comparative_study}--\ref{subsec:portability_analysis}.
\item Appendix~\ref{subsec:details_image_generation_128}:
  Implementation details of $128 \times 128$ image generation, which were used to generate samples in Figure~\ref{fig:concept}.
\item Appendix~\ref{subsec:details_shallow_dof_rendering}:
  Implementation details of shallow DoF rendering, which were used in the experiments in Section~\ref{subsec:application}.
\end{itemize}

\smallskip\noindent\textbf{Notation.}
In our description of the network architectures, we use the following notation.
\begin{itemize}
  \setlength{\parskip}{1pt}
  \setlength{\itemsep}{1pt}
\item Linear:
  Linear layer
\item Conv:
  Convolutional layer
\item Deconv:
  Deconvolutional (i.e., fractionally strided convolutional) layer
\item ReLU:
  Rectified linear unit~\cite{VNairICML2010}
\item LReLU:
  Leaky rectified linear unit~\cite{AMaasICML2013,BXuICMLW2015}
\item IN:
  Instance normalization~\cite{DUlyanovArXiv2016}
\item AdaIN:
  Adaptive instance normalization~\cite{XHuangICCV2017}
\item SN:
  Spectral normalization~\cite{TMiyatoICLR2018}
\end{itemize}

In our description of the training settings, we use the following notation.
We used the Adam optimizer~\cite{DPKingmaICLR2015} for training.
\begin{itemize}
  \setlength{\parskip}{1pt}
  \setlength{\itemsep}{1pt}
\item $\alpha$:
  Learning rate
\item $\beta_{1}$:
  First-order momentum parameter
\item $\beta_{2}$:
  Second-order momentum parameter
\end{itemize}

\subsection{Details of image generation (Sections~\ref{subsec:comparative_study}--\ref{subsec:portability_analysis})}
\label{subsec:details_image_generation}

\begin{table}
  \centering
  \footnotesize
  \begin{subtable}{0.85\columnwidth}
    \caption{Generators $G_I$ and $G_D$}
    \vspace{-2mm}
    \begin{tabularx}{\columnwidth}{C}
      \toprule
      Input 1: Constant $c \in \mathbb{R}^{4 \times 4 \times 1024}$,
      \\
      Input 2: Random noise $z \in \mathbb{R}^{128} \sim {\cal N}(0, 1)$
      \\ \midrule
      $c$, AdaIN(z), ReLU
      \\ \midrule
      $4 \times 4$ Deconv up 512, AdaIN(z), ReLU
      \\ \midrule
      $4 \times 4$ Deconv up 256, AdaIN(z), ReLU
      \\ \midrule
      $4 \times 4$ Deconv up 128, AdaIN(z), ReLU
      \\ \midrule
      $4 \times 4$ Deconv up 64, AdaIN(z), ReLU
      \\ \midrule
      $4 \times 4$ Conv 3, Tanh $\rightarrow$ $I_d^g$
      \\
      $4 \times 4$ Conv 1, Tanh $\times 10 \times \text{MLP}(z)$ $\rightarrow$ $D^g$
      \\ \bottomrule
    \end{tabularx}
  \end{subtable}
  \begin{subtable}{0.85\columnwidth}
    \vspace{2mm}
    \caption{Depth expansion network $T$}
    \vspace{-2mm}
    \begin{tabularx}{\columnwidth}{C}
      \toprule
      Input: Warped depth map $D(\mathbf{x} + \mathbf{u}D(\mathbf{x})) \in \mathbb{R}^{64 \times 64 \times 25}$
      \\ \midrule
      $3 \times 3$ Conv 25, IN, LReLU
      \\ \midrule
      $3 \times 3$ Conv 25, IN, LReLU
      \\ \midrule
      $3 \times 3$ Conv 25, IN, LReLU, Add input $\rightarrow$ $M(\mathbf{x}, \mathbf{u})$
      \\ \bottomrule
    \end{tabularx}
  \end{subtable}
  \begin{subtable}{0.85\columnwidth}
    \vspace{2mm}
    \caption{Discriminator $C$}
    \vspace{-2mm}
    \begin{tabularx}{\columnwidth}{C}
      \toprule
      Input: Image $I \in \mathbb{R}^{64 \times 64 \times 3}$
      \\ \midrule
      $5 \times 5$ Conv down 64, LReLU
      \\ \midrule
      $5 \times 5$ SN Conv down 128, IN, LReLU
      \\ \midrule
      $5 \times 5$ SN Conv down 256, IN, LReLU
      \\ \midrule
      $5 \times 5$ SN Conv down 512, IN, LReLU
      \\ \midrule
      Linear
      \\ \bottomrule
    \end{tabularx}
  \end{subtable}
  \vspace{1.5mm}
  \caption{\textbf{AR-GAN architecture for $64 \times 64$ images.}}
  \label{tab:argan64}
  \vspace{-2mm}
\end{table}

\noindent\textbf{Network architectures.}
Table~\ref{tab:argan64} provides the AR-GAN architecture for $64 \times 64$ images.
This was used in the experiments in Sections~\ref{subsec:comparative_study}--\ref{subsec:portability_analysis}.
As mentioned in Section~\ref{sec:experiments}, we designed the generator (Table~\ref{tab:argan64}(a)) and discriminator (Table~\ref{tab:argan64}(c)) based on HoloGAN~\cite{TNguyenICCV2019},\footnote{\label{foot:hologan}\url{https://github.com/thunguyenphuoc/HoloGAN}} which was used as a baseline in the experiments.
In HoloGAN, 3D transformation and 3D convolution were used in the generator.
However, they are not required in AR-GANs.
Therefore, in AR-GAN, we removed the 3D transformation and replaced the 3D convolution with a 2D convolution.

The generator (Table~\ref{tab:argan64}(a)) has a StyleGAN-like architecture~\cite{TKarrasCVPR2019}, in which the latent vector $z$ is inserted into the network using adaptive instance normalization ($\text{AdaIN}(z)$)~\cite{XHuangICCV2017}.
In the last layer of the generator, we adjusted the scale of $D$ in an instance-dependent manner using a constant (i.e., 10) and a one-hidden-layer multilayer perceptron ($\text{MLP}(z)$).
Table~\ref{tab:argan64}(b) provides the architecture of the depth expansion network $T$, which was used in the aperture renderer (Section~\ref{subsec:lf_aperture_rendering}).
In the aperture renderer, we set the aperture size to $5 \times 5$.
This was implemented based on a previous study on aperture rendering~\cite{PPSrinivasanCVPR2018}.
In the discriminator (Table~\ref{tab:argan64}(c)), we used instance normalization (IN)~\cite{DUlyanovArXiv2016} and spectral normalization (SN)~\cite{TMiyatoICLR2018}, following the implementation of HoloGAN.\footnoteref{foot:hologan}

We implemented HoloGAN (used in Section~\ref{subsec:comparative_study}) using the same network architecture that was used in the original study~\cite{TNguyenICCV2019}.\footnoteref{foot:hologan}
In the generator, the first and second 2D deconvolutional layers in Table~\ref{tab:argan64}(a) were replaced with 3D deconvolutional layers.
After this processing, 3D transformation was performed, and then $1 \times 1$ Conv and ReLU were applied to adjust the number of channels.
Strictly, we modified the number of channels following the official code.\footnoteref{foot:hologan}
In addition, we removed a depth estimation layer (which is presented at the bottom of Table~\ref{tab:argan64}(a)) because HoloGAN cannot optimize it.
In HoloGAN, we used the same discriminator as the one used in AR-GAN (Table~\ref{tab:argan64}(c)).

We implemented the generator and discriminator in RGBD-GAN (used in Section~\ref{subsec:comparative_study}) using almost the same network architecture as that of AR-GAN (Table~\ref{tab:argan64}).
An exception is that we conducted a modification to the generator (Table~\ref{tab:argan64}(a)) to incorporate the viewpoint information (i.e., azimuth $\theta_a$ and elevation $\theta_e$) into the network.
More concretely, we concatenated $c_{cyclic} = [\cos(\theta_a), \sin(\theta_a), \cos(\theta_e), \sin(\theta_e)]$ to $z$ and used the combination of $z$ and $c_{cyclic}$ as the input of AdaIN in the first and second layers.
In the initial experiments, we found that replacing the constant input $c$ with $c_{cyclic}$ (by converting $c_{cyclic}$ using a CNN to match the size) is useful for learning the disentangled representations.
Therefore, we applied this modification in the experiments.
In addition, we modified the depth estimation layer at the bottom of Table~\ref{tab:argan64}(a) as ``$4 \times 4$ Conv 1, Tanh, Shift 1'' because we found that this setting works reasonably well.
In RGBD-GAN, we used the same discriminator as that used in AR-GAN (Table~\ref{tab:argan64}(c)).

When implementing AR-HoloGAN/AR-RGBD-GAN (used in Section~\ref{subsec:portability_analysis}), we added a depth estimation layer (shown at the bottom of Table~\ref{tab:argan64}(a)) to the last layer of the HoloGAN/RGBD-GAN generator.
Through this modification, we generated $I_d^g$ and $D^g$ simultaneously, and then rendered the shallow DoF image (i.e., $I_s^g$) from $(I_d^g, D^g)$ using the aperture renderer, similar to AR-GAN.
In addition, in AR-HoloGAN and AR-RGBD-GAN, we used the same discriminator as that used in AR-GAN (Table~\ref{tab:argan64}(c)).

\smallskip\noindent\textbf{Training settings.}
As a GAN objective function, we used a non-saturating GAN loss~\cite{IGoodfellowNIPS2014}.
We set the hyperparameters of the center focus prior (Equation~\ref{eqn:center_focus_prior}) as $r_{\text{th}} = 0.25$ and $g = 1$, where we scale $r$ (the distance from the center of the image) such that $r = 1$ represents half of the image width or height.
We set the weighting parameter $\lambda_p$ (Equation~\ref{eqn:prior}) to 1.
We used the center focus prior only during the initial $5k$, $5k$, and $50k$ iterations on Oxford Flowers, CUB-200-2011, and FFHQ, respectively.
Following a previous study~\cite{PPSrinivasanCVPR2018}, we regularized the depth expansion network $T$ so that the predicted depth map $M(\mathbf{x}, \mathbf{u})$ was close to the warped versions of $D(\mathbf{x})$:
\begin{flalign}
  \label{eqn:regularization_depth_map}
  {\cal L}_d = \lambda_d \| M(\mathbf{x}, \mathbf{u}) - D(\mathbf{x} + \mathbf{u}D(\mathbf{x})) \|_1,
\end{flalign}
where $\lambda_d$ is the weighting parameter and set to 1 in the experiments.

We trained the networks from scratch using the Adam optimizer~\cite{DPKingmaICLR2015} with $\alpha = 0.0001$, $\beta_1 = 0.5$, $\beta_2 = 0.999$, and a batch size of 32.
Following the implementation of HoloGAN, we updated the generator twice for every update of the discriminator.
We updated the discriminator for $150k$, $250k$, and $350k$ iterations on Oxford Flowers, CUB-200-2011, and FFHQ, respectively.
To stabilize the training, we applied differentiable augmentation~\cite{SZhaoNeurIPS2020}, including color jittering, translation, and cutout~\cite{TDevriesArXiv2017}.
To obtain a stable performance, we used an exponential moving average~\cite{TKarrasICLR2017} with a decay of 0.999 over the weights to produce the final generator.

When training HoloGAN, RGBD-GAN, AR-HoloGAN, and AR-RGBD-GAN, we sampled the azimuth and elevation uniformly within the range of $[-50, 50]$ and $[-20, 20]$, respectively, because such range values were used as default values in HoloGAN and RGBD-GAN (particularly, on a face image dataset).

\subsection{Details of image generation (Figure~\ref{fig:concept})}
\label{subsec:details_image_generation_128}

\begin{table}
  \centering
  \footnotesize
  \begin{subtable}{0.85\columnwidth}
    \caption{Generators $G_I$ and $G_D$}
    \vspace{-2mm}
    \begin{tabularx}{\columnwidth}{C}
      \toprule
      Input 1: Constant $c \in \mathbb{R}^{4 \times 4 \times 1024}$,
      \\
      Input 2: Random noise $z \in \mathbb{R}^{128} \sim {\cal N}(0, 1)$
      \\ \midrule
      $c$, AdaIN(z), ReLU
      \\ \midrule
      $4 \times 4$ Deconv up 1024, AdaIN(z), ReLU
      \\ \midrule
      $4 \times 4$ Deconv up 512, AdaIN(z), ReLU
      \\ \midrule
      $4 \times 4$ Deconv up 256, AdaIN(z), ReLU
      \\ \midrule
      $4 \times 4$ Deconv up 128, AdaIN(z), ReLU
      \\ \midrule
      $4 \times 4$ Deconv up 64, AdaIN(z), ReLU
      \\ \midrule
      $4 \times 4$ Conv 3, Tanh $\rightarrow$ $I_d^g$
      \\
      $4 \times 4$ Conv 1, Tanh $\times 10 \times \text{MLP}(z)$ $\rightarrow$ $D^g$
      \\ \bottomrule
    \end{tabularx}
  \end{subtable}
  \begin{subtable}{0.85\columnwidth}
    \vspace{2mm}
    \caption{Depth expansion network $T$}
    \vspace{-2mm}
    \begin{tabularx}{\columnwidth}{C}
      \toprule
      Input: Warped depth map $D(\mathbf{x} + \mathbf{u}D(\mathbf{x})) \in \mathbb{R}^{128 \times 128 \times 25}$
      \\ \midrule
      $3 \times 3$ Conv 25, IN, LReLU
      \\ \midrule
      $3 \times 3$ Conv 25, IN, LReLU
      \\ \midrule
      $3 \times 3$ Conv 25, IN, LReLU, Add input $\rightarrow$ $M(\mathbf{x}, \mathbf{u})$
      \\ \bottomrule
    \end{tabularx}
  \end{subtable}
  \begin{subtable}{0.85\columnwidth}
    \vspace{2mm}
    \caption{Discriminator $C$}
    \vspace{-2mm}
    \begin{tabularx}{\columnwidth}{C}
      \toprule
      Input: Image $I \in \mathbb{R}^{128 \times 128 \times 3}$
      \\ \midrule
      $5 \times 5$ Conv down 64, LReLU
      \\ \midrule
      $5 \times 5$ SN Conv down 128, IN, LReLU
      \\ \midrule
      $5 \times 5$ SN Conv down 256, IN, LReLU
      \\ \midrule
      $5 \times 5$ SN Conv down 512, IN, LReLU
      \\ \midrule
      Linear
      \\ \bottomrule
    \end{tabularx}
  \end{subtable}
  \vspace{1.5mm}
  \caption{\textbf{AR-GAN architecture for $128 \times 128$ images.}}
  \label{tab:argan128}
  \vspace{-2mm}
\end{table}

\noindent\textbf{Network architectures.}
Table~\ref{tab:argan128} provides the AR-GAN architecture for $128 \times 128$ images.
This was used for generating samples in Figure~\ref{fig:concept}.
This network architecture is similar to that for $64 \times 64$ images (Table~\ref{tab:argan64}) except that one layer is added to the generator according to the change in image size.

\smallskip\noindent\textbf{Training settings.}
We used similar training settings as those for $64 \times 64$ images (Section~\ref{subsec:details_image_generation}) except that the hyperparameters of the center focus prior were modified as $r_{\text{th}} = 0.25$ and $g = 2$ according to the change in image size.
We used the center focus prior only during the initial $5k$, $5k$, and $5k$ iterations on Oxford Flowers, CUB-200-2011, and FFHQ, respectively.

\subsection{Details of shallow DoF rendering (Section~\ref{subsec:application})}
\label{subsec:details_shallow_dof_rendering}

\begin{table}[tb]
  \centering
  \footnotesize
  \begin{tabularx}{0.85\columnwidth}{cC}
    \toprule
    \multicolumn{2}{c}{Input: $I_d \in \mathbb{R}^{128 \times 128 \times 3}$}
    \\ \midrule
    \textsc{enc{\_}conv0} & $3 \times 3$ Conv $48$, LReLU
    \\
    \textsc{enc{\_}conv1} & $3 \times 3$ Conv $48$, LReLU
    \\
    \textsc{pool1} & $2 \times 2$ Maxpool
    \\ \midrule
    \textsc{enc{\_}conv2} & $3 \times 3$ Conv $48$, LReLU
    \\
    \textsc{pool2} & $2 \times 2$ Maxpool
    \\ \midrule
    \textsc{enc{\_}conv3} & $3 \times 3$ Conv $48$, LReLU
    \\
    \textsc{pool3} & $2 \times 2$ Maxpool
    \\ \midrule
    \textsc{enc{\_}conv4} & $3 \times 3$ Conv $48$, LReLU
    \\
    \textsc{pool4} & $2 \times 2$ Maxpool
    \\ \midrule
    \textsc{enc{\_}conv5} & $3 \times 3$ Conv $48$, LReLU
    \\
    \textsc{pool5} & $2 \times 2$ Maxpool
    \\ \midrule
    \textsc{enc{\_}conv6} & $3 \times 3$ Conv $48$, LReLU
    \\ \midrule
    \textsc{upsample5} & $2 \times 2$ Upsample
    \\
    \textsc{concat5} & Concatenate output of \textsc{pool4}
    \\
    \textsc{dec{\_}conv5a} & $3 \times 3$ Conv $96$, LReLU
    \\
    \textsc{dec{\_}conv5b} & $3 \times 3$ Conv $96$, LReLU
    \\ \midrule
    \textsc{upsample4} & $2 \times 2$ Upsample
    \\
    \textsc{concat4} & Concatenate output of \textsc{pool3}
    \\
    \textsc{dec{\_}conv4a} & $3 \times 3$ Conv $96$, LReLU
    \\
    \textsc{dec{\_}conv4b} & $3 \times 3$ Conv $96$, LReLU
    \\ \midrule
    \textsc{upsample3} & $2 \times 2$ Upsample
    \\
    \textsc{concat3} & Concatenate output of \textsc{pool2}
    \\
    \textsc{dec{\_}conv3a} & $3 \times 3$ Conv $96$, LReLU
    \\
    \textsc{dec{\_}conv3b} & $3 \times 3$ Conv $96$, LReLU
    \\ \midrule
    \textsc{upsample2} & $2 \times 2$ Upsample
    \\
    \textsc{concat2} & Concatenate output of \textsc{pool1}
    \\
    \textsc{dec{\_}conv2a} & $3 \times 3$ Conv $96$, LReLU
    \\
    \textsc{dec{\_}conv2b} & $3 \times 3$ Conv $96$, LReLU
    \\ \midrule
    \textsc{upsample1} & $2 \times 2$ Upsample
    \\
    \textsc{concat1} & Concatenate output of \textsc{input}
    \\
    \textsc{dec{\_}conv1a} & $3 \times 3$ Conv $64$, LReLU
    \\
    \textsc{dec{\_}conv1b} & $3 \times 3$ Conv $32$, LReLU
    \\
    \:\:\textsc{dec{\_}conv1c}\:\: & $3 \times 3$ Conv $1$ $\rightarrow$ $D$
    \\ \bottomrule
  \end{tabularx}
  \vspace{1.5mm}
  \caption{\textbf{AR-GAN-DR depth estimator architecture.}}
  \vspace{-2mm}
  \label{tab:depth_estimator}
\end{table}

\noindent\textbf{Network architectures.}
Table~\ref{tab:depth_estimator} provides the AR-GAN-DR depth estimator architecture.
We used the u-net architecture~\cite{ORonnebergerMICCAI2015}.
The AR-GAN-R image translator has a similar architecture to this, except that the number of channels of the last layer is modified to three to generate an RGB image.
These models were used in the experiments in Section~\ref{subsec:application}.

\smallskip\noindent\textbf{Training settings.}
We generated training data (i.e., pairs of images and depths) using the AR-GAN that was trained using $128 \times 128$ images on Oxford Flowers.
This AR-GAN was the same as that used for generating the samples in Figure~\ref{fig:concept}.
We trained the depth estimator for $300k$ iterations using the Adam optimizer~\cite{DPKingmaICLR2015} with $\alpha = 0.0003$, $\beta_1 = 0.9$, $\beta_2 = 0.99$, and a batch size of 4.
The learning rate was kept constant during the training, except for the last $30\%$ iterations, where the learning rate was smoothly ramped down to zero.

\end{document}